\documentclass[a4paper,final,12pt,onecolumn]{IEEEtran}
\usepackage{setspace}
   \usepackage[dvips]{graphicx}
   \graphicspath{{../eps/}}
   \DeclareGraphicsExtensions{.pdf,.jpeg,.png,.eps}
\usepackage[cmex10]{amsmath}
\usepackage{amsthm}
\theoremstyle{definition}
\newtheorem{Definition}{Definition}[section]

\usepackage[tight,footnotesize]{subfigure}

\begin{document}
%
\title{Efficient neuro-fuzzy system and its Memristor Crossbar-based Hardware Implementation}

%
%
%

\author{Farnood~Merrikh-Bayat and
        Saeed Bagheri-Shouraki}
\maketitle

\begin{abstract}
In this paper a novel neuro-fuzzy system is proposed where its
learning is based on the creation of fuzzy relations by using new
implication method without utilizing any exact mathematical
techniques. Then, a simple memristor crossbar-based analog circuit
is designed to implement this neuro-fuzzy system which offers very
interesting properties. In addition to high connectivity between
neurons and being fault-tolerant, all synaptic weights in our
proposed method are always non-negative and there is no need to
precisely adjust them. Finally, this structure is hierarchically
expandable and can compute operations in real time since it is
implemented through analog circuits. Simulation results show the
efficiency and applicability of our neuro-fuzzy computing system.
They also indicate that this system can be a good candidate to be
used for creating artificial brain.
\end{abstract}

\section{introduction}
In the field of artificial intelligence, neuro-fuzzy refers to
combination of artificial neural networks and fuzzy logics trying to
use benefits of these two fields. Perhaps the most important
advantage of neural networks is their adaptivity. Adaptivity comes
from learning capability of neural networks which allows these
networks to perform well even when the environment varies over time
like what human brain does. Another significant benefit of neural
networks relates to their huge connectivity (again inspired from
real brain) that offers high fault tolerance and parallel processing
power which is also assumed to be the reason of high efficiency of
biological systems. On the other hand, fuzzy logic performs an
inference mechanism under cognitive uncertainty and provides an
inference morphology that enables approximate human reasoning
capabilities to be applied to knowledge-based systems. However, in
fuzzy inference systems, it usually takes a lot of time to design
and tune membership functions and rules. To overcome this problem,
similar to what is done in neuro-fuzzy systems, learning techniques
of neural networks can automated this tuning process to reduce
development time.

Therefore, it can be said that current neuro-fuzzy systems like
ANFIS \cite{anfis}, RuleNet \cite{RulNet} and GARIC \cite{GARIC} are
fuzzy systems that use a learning algorithm derived from neural
network theory to determine their parameters (fuzzy sets and fuzzy
rules) by processing training data. However, eventually all of these
neuro-fuzzy systems are trying to approach an ideal soft-computing
tool where the nature of its computing or inference be as similar as
possible to computation and inference in human brain. Although now
we can see significant progresses in this area in the software
domain, samples of successful hardware which could even approach
computing capabilities of real brain are very rare. One of the main
obstacle in front of this purpose relates to the ability of
efficiently modeling and construction of synapses. Actually, highly
parallel processing power of biological systems comes from large
connectivity between neurons through synapses (each neuron in human
brain is connected to about 10000 other neurons through these
synapses) and it is widely believed that the adaptation of synaptic
weights enables the biological systems to learn and function.
Therefore, efficient construction of synaptic weights is a critical
factor in the success of final system. After the first physical
realization of memristor \cite{williams}, it becomes clear that this
passive element can be a perfect representative of synapse since
similar to synapse, its conductance can be precisely modulated by
passing charge and flux through it.

In the past few years and by the discovery of memristor and
improvements achieved in the construction of powerful digital
hardware, extensive works are in progress to build an artificial
brain that can adaptively interact with the world in real time.
Among then we can name projects like IFAT 4G at John Hopkins
University, BrainScales in the European Union's neuromorphic chip
program, brain simulator C2 introduced by IBM, and Modular Neural
Exploring Traveling Agent (MoNETA). The most recent one between
these projects, {\it i.e.} MoNETA which is the part of the DARPA's
SyNAPSE program, is a software developed by the researchers at
Boston University which will run on a brain-inspired microprocessor
under development at HP labs in California \cite{Moneta}. In this
still under construction system, in spite of internal structure of
real brain, all units are implemented in digital with separate
memory and computational units. However, it has been argued that by
constructing memristive memories (to store synaptic weights) and
putting them very close to computational circuits that read and
write them, signalling losses and power consumption can be minimized
\cite{Moneta}.

In this paper, we introduce another approach to construct a simple
neuro-fuzzy computing system which differs significantly from other
systems in this category. In fact, we believe that structures
similar to what is proposed in this paper merit more to be called
neuro-fuzzy compared to currently available systems. This is because
of the fact that as can be seen in the rest of the paper, in our
proposed system neural network and fuzzy logic fields are completely
involved with each other and it is hard to distinguish that which
part belongs to fuzzy logic and which part relates to neural
networks. Moreover, there is no use of exact mathematical
techniques. As another advantage, in this system memristor crossbar
is used to store synaptic weights but by this difference that memory
is assimilated with computational units like what we have in human
brain. In addition, because of large connectivity between input and
output neurons, this proposed structure is completely
fault-tolerant. Actually we have shown that even if during the
fabrication process or execution phase of the system near half of
the memristors become faulty, system can still continue working
satisfactory. On the other hand, learning in our structure is based
on the creation of fuzzy relations which is shown to be equal to
primary Hebbian learning rule and therefore all synaptic weights
will be always non-negative. However, in spite of almost all
developed learning methods, there is no need to precisely adjust
these weights which are represented by memristors. Finally, our
neuro-fuzzy system is hierarchically expandable and since it is
constructed with analog circuits, its computations are almost in
real-time.

The paper is organized as follows. Inference method in our proposed
neuro-fuzzy system is described in Section \ref{neurofuzzy}. Section
\ref{hardware} is devoted to the explanation of the hardware
implementation of proposed inference method based on memristor
crossbar structure. The reason that we have considered our proposed
system as a neuro-fuzzy one is presented in Section \ref{reason}.
Eventually, a few experimental results are presented in Section
\ref{simulation}, before conclusions in Section \ref{conclusion}.

\section{construction of a new fuzzy neuro-fuzzy structure}
\label{neurofuzzy} 
\begin{figure}[!t]
\centering \subfigure[]{
\label{figa:1} 
\includegraphics[width=5in,height=1.2in]{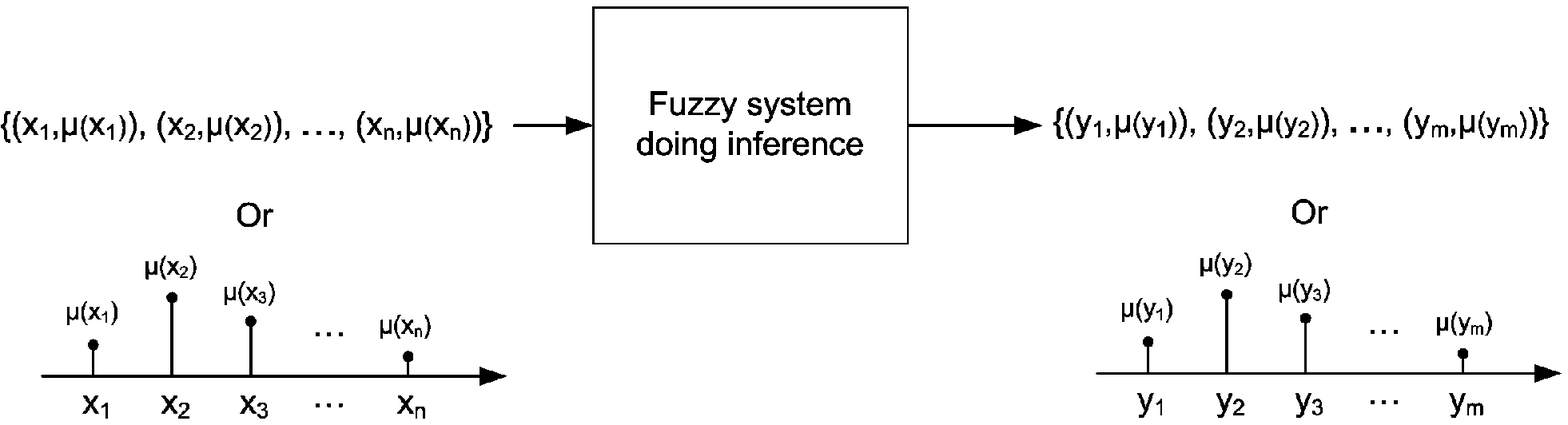}}
\hspace{0.08in} \subfigure[]{
\label{figa:2} 
\includegraphics[width=2.5in,height=1.3in]{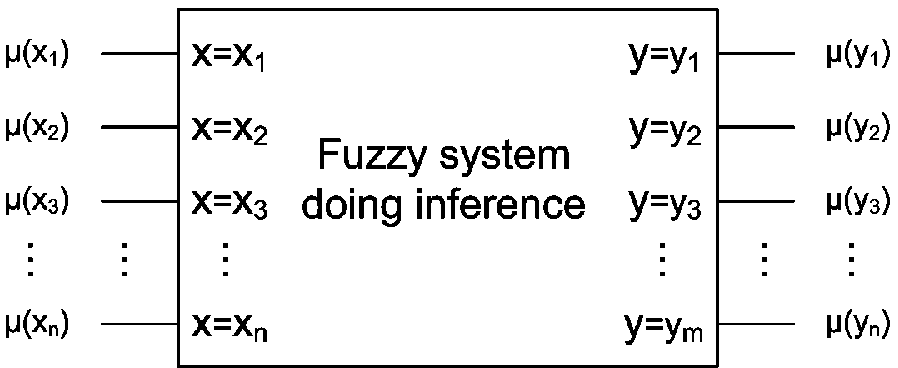}}
\vspace{0.22in} \subfigure[]{
\label{figa:3} 
\includegraphics[width=2.5in,height=1.3in]{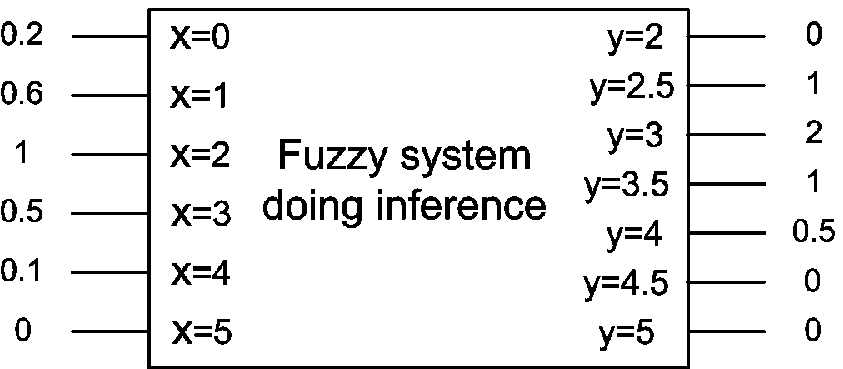}}
\hspace{0.01in} \subfigure[]{
\label{figa:4} 
\includegraphics[width=3.5in,height=2.3in]{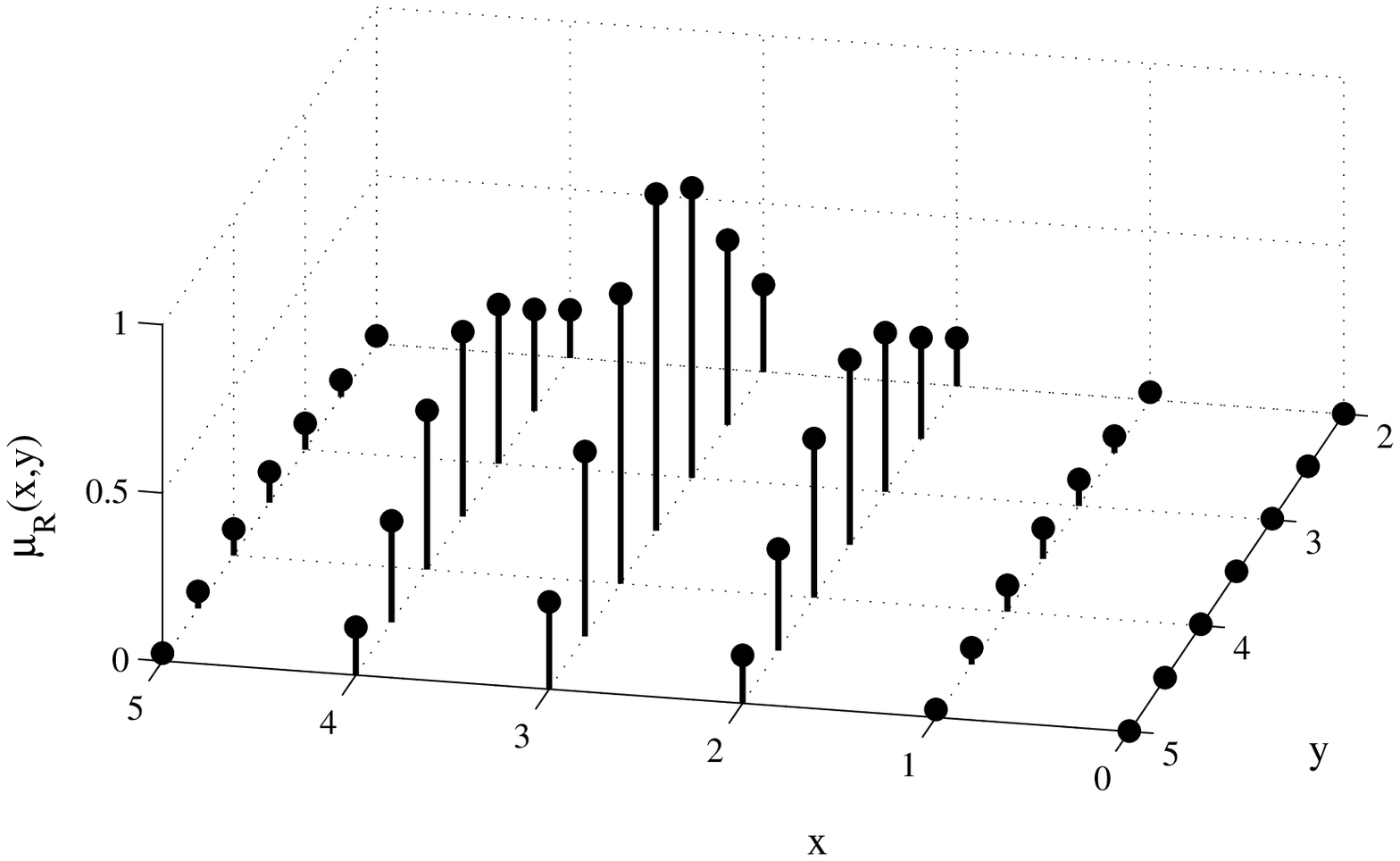}}
 \caption{Construction of fuzzy inference system. (a) A typical discrete fuzzy inference system with fuzzy input and output variables.
 (b) Fuzzy inference system with modified input and output terminals where its hardware implementation is much simpler than the system of Fig. \ref{figa:1}.
 (c) A typical example of the structure of Fig. \ref{figa:2}. (d) A typical fuzzy relation for the system of Fig. \ref{figa:3} where fuzzy
 reasoning can be done based on it and given fuzzy number.}
\label{figa} 
\end{figure}

In this paper we propose a new fuzzy structure
with biological support which can do some inferential tasks in fuzzy
form. In addition, this fuzzy structure has learning capability like
human brain but in spite of neural networks, it performs any
operation based on concepts of fuzzy logic. This means that in our
proposed structure, inputs and outputs of the structure are fuzzy
numbers and any calculation in the structure for doing inferential
processes is done in fuzzy without using any precise and accurate
mathematical techniques. Figure \ref{figa:1} shows such a typical
Single Input Single Output (SISO) system where in this system, fuzzy
input and output variables are considered to be discrete for
simplicity. If this system was a non-fuzzy system, its input or
output would be a single crisp number. However, since this system is
a fuzzy one, its input and output should be fuzzy numbers or set of
pairs like what is depicted in Fig. \ref{figa:1}. Since the
construction of this system with this kind of input and output
terminals is impossible (from hardware aspect), we modify input and
output terminals of the fuzzy system of Fig.\ref{figa:1} such as the
one presented in Fig. \ref{figa:2}. In this figure, the {\it i}th
input and the {\it j}th output of the system represent concepts
$x=x_i$ and $y=y_j$ respectively. Note that resolution or domain of
input or output variables can be increased by adding input or output
terminals to the system where each of these newly added terminals
will represent new values of input or output variables. Now, if in
the system of Fig. \ref{figa:2}, the input applied to the {\it i}th
input terminal which represents concept $x=x_i$ be $\mu_{A'}(x=x_i)$
($\mu_{A'}$: membership function of input fuzzy number) and the
output at the {\it j}th output terminal which represents concept
$y=y_j$ be $\mu_{B'}(y=y_j)$ ($\mu_{B'}$: membership function of
output fuzzy number), the system of Fig. \ref{figa:2} will be equal
to the fuzzy system of Fig. \ref{figa:1} but by this difference that
the construction of the fuzzy system shown in Fig. \ref{figa:2} is
much simpler. Consequently, input of the system of Fig. \ref{figa:2}
is a vector
 of membership degrees, {\it i.e.} $\left[\mu(x=x_1), \mu(x=x_2), \ldots,
 \mu(x=x_n)\right]$, and the output of this system is also a vector
 of membership degrees, {\it i.e.} $\left[\mu(y=y_1), \mu(y=y_2), \ldots,
 \mu(y=y_m)\right]$ where combination of these membership grades with those concepts which are assigned
 to input and output terminals creates input and output fuzzy numbers. As a result, we again have a totally fuzzy system
 with fuzzy input and fuzzy output variables. For example, consider a typical fuzzy system shown in Fig. \ref{figa:3}. In
 this example, input of the system is a fuzzy number
 $\{(0,0.2),(1,0.6),(2,1),(3,0.5),(4,0.1),(5,0)\}$ which can be
 roughly considered as a fuzzy representation of crisp number $x=2$ and the output of this system for this input is a fuzzy number
 $\{(2,0),(2.5,1),(3,2),(3.5,1),(4,0.5),(4.5,0),(5,0)\}$ where its deffuzzification
 may probably result in a crisp number $y=4$.

 Now, let's see how the fuzzy inference system of Fig. \ref{figa:2}
 which relates the fuzzy input variable to fuzzy output variable can be constructed.
 It is well known that the compositional rule of inference describes a composition of a fuzzy set and a fuzzy relation. Any fuzzy
 rule in the form of
\begin{equation}\label{eqinf}
\text{IF}\quad x\quad \text{is}\quad  A\quad  \text{THEN}\quad
y\quad \text{is}\quad B
\end{equation}
is usually represented by a fuzzy relation $R$ \cite{leszek}. Having
given input linguistic value $A'$, we can infer an output fuzzy set
$B'$ by the composition of the fuzzy set $A'$ and the relation $R$.
Therefore, doing any inference between fuzzy input and output
 variables requires a fuzzy relation $R$ between these variables.
 Each fuzzy relation can be represented by a surface which we call it
{\it Fuzzy Relation Surface} or FRS. Value of this surface at any
point is from a kind of membership degree and therefore
 will be always non-negative. In literature, several implication methods have been proposed
 to construct a fuzzy relation based on given input and output fuzzy
 sets \cite{zadehimplication,fodor,dubois}. In this paper, we will show another method to construct
 fuzzy relation based on available training fuzzy data when input and output fuzzy sets are not
 known. As can be seen in the rest of this paper, our proposed
 method has the ability of learning as well.

Figure \ref{figa:4} shows one typical FRS for the fuzzy system of
Fig. \ref{figa:3}. Note that since input and output variables of the
fuzzy system of Fig. \ref{figa:3} are discrete, this surface has
became discrete as well. One reason that we have made input and
output of this system and consequently its representing fuzzy
relation discrete is that the hardware implementation of continues
fuzzy system is very difficult if not impossible. As stated before,
it is evident that based on the requirements of the application that
this fuzzy system is intended for, resolution and range of its
corresponding discrete fuzzy relation (surface) can be increased
simply by increasing the number of input and(or) output terminals of
the system.

In its simplest form, any fuzzy relation can be defined as follows
\cite{leszek}:

\begin{Definition}[Fuzzy Relation]
Let $\mathbf{X}$ and $\mathbf{Y}$ be two universes of discourse.
Binary fuzzy relation, denoted by $R$, are fuzzy sets which map each
element in the product set $\mathbf{X}\times\mathbf{Y}$ to a
membership grade $\mu_R(x,y)$, where $x\in \mathbf{X}$ and $y\in
\mathbf{Y}$. So any fuzzy relation can be expressed as:
\begin{equation}
R=\{\left((x,y), \mu_R(x,y)\right)| \mu_R(x,y)\geq 0,\ x\in
\mathbf{X},\ y\in \mathbf{Y}\}
\end{equation}
\end{Definition}

Now the problem is to determine the membership function of the fuzzy
relation $R$ based on given input and output fuzzy sets. In this
section, two different situation will be considered: (i) input and
output fuzzy sets are available and (ii) input and output fuzzy sets
are not in hand but instead of them, some fuzzy input-output
training data are available.

\subsection{construction of fuzzy relation based on available input
and output fuzzy sets} Consider that input fuzzy set $A$ with
membership function $\mu_A$ and output fuzzy set $B$ with membership
function $\mu_B$ are available and the goal is to determine the
membership function of the fuzzy relation described by:
\begin{equation}
\mu_R(x,y)=\mu_{A\rightarrow B}=I(\mu_A(x), \mu_B(y)),\ \forall x\in
X\ \text{and}\ y\in Y,
\end{equation}
based on the knowledge of $\mu_A(x)$ and $\mu_B(y)$ where $I$ is a
fuzzy implication scheme. Here, we consider a new and simple
implication method defined as:
\begin{equation}\label{eqfr}
\mu_R(x,y)=I(\mu_A(x), \mu_B(y))=f(\mu_A(x)+\mu_B(y)),\ \forall x\in
X\ \text{and}\ y\in Y,
\end{equation}
where $f(\cdot,\cdot)$ can be any monolithically increasing
function. Note that since we are dealing with fuzzy systems,
function $f(\cdot,\cdot)$ does not need to be defined precisely.
Equation \ref{eqfr} says that the membership value of the
constructed fuzzy relation at any point $(x, y)$ should be directly
proportional to the sum of the values of the input's  membership
function at point $x$ and the output's membership function at point
$y$. In other words, whatever both values of input and output
membership functions at points $x$ and $y$ respectively be higher,
the membership value of fuzzy relation $R$ at point $(x, y)$ should
be higher as well. Although this implication method is not similar
to other common methods, it has this benefit that as can be seen
later, its hardware implementation is straightforward.

\subsection{construction of fuzzy relation based on available fuzzy
training data} \label{inkdrop}

Now, consider the case in which input and output fuzzy sets are not
available but we want to construct a fuzzy relation based on $N$
existing fuzzy input-output training data. Each fuzzy input-output
training data for any SISO system consists of two fuzzy numbers:
input fuzzy number denoted by $A'$ and output fuzzy number denoted
by $B'$. Therefore, the {\it i}th fuzzy training data can be
considered as $\{A'_i,B'_i\}$ where:
\begin{eqnarray}
A'_i&=&\{(x,\mu_{A'_i}(x))|\  x\in \mathbf{X}\}, \ \forall
x\in\mathbf{X},\nonumber\\
B'_i&=&\{(y,\mu_{B'_i}(y))| \ y\in \mathbf{Y}\},\  \forall
y\in\mathbf{Y},
\end{eqnarray}

 In above equations, $\mathbf{X}$ and $\mathbf{Y}$ are universes of discourse and $\mu_{A'_i}$ and $\mu_{B'_i}$
 are membership functions of fuzzy numbers $A'_i$ and $B'_i$ respectively. Similar to what is done in previous subsection, we propose to construct a fuzzy relation representing the relation between input and output fuzzy variables as:
\begin{equation}
\mu_R(x,y)=\sum_{i=1}^N f(\mu_{A'_i}(x)+\mu_{B'_i}(y)),\ \forall
x\in \mathbf{X}\ \text{and}\ y\in \mathbf{Y}.
\end{equation}

\begin{figure}[!t]
\centering 
{
\includegraphics[width=3.2in,height=2.6in]{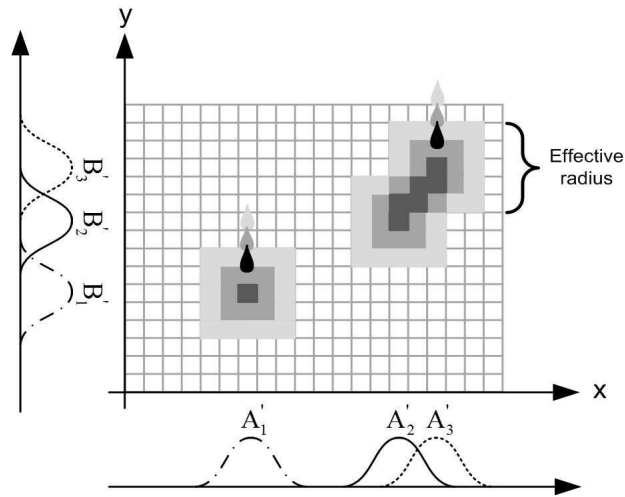}}
\caption{Ink drop spreading process used to construct fuzzy relation
based on available fuzzy training data.}
\label{figfuzrelcreation} 
\end{figure}

However, when all training data are not entirely available at the
beginning or when they are entering one by one, fuzzy relation can
be formed iteratively. In this case, entrance of any new data should
update the current fuzzy relation which its updating rule can be
written as:
\begin{equation}\label{updateink}
\mu_R^{new}(x,y)=
\mu_R^{old}(x,y)+f(\mu_{A'_{new}}(x)+\mu_{B'_{new}}(y)), \ \forall
x\in \mathbf{X}\ \text{and} \ y\in \mathbf{Y}.
\end{equation}
where $\{A'_{new},B'_{new}\}$ is the newly observed fuzzy
input-output training data. Here it is worth to discuss a bit more
about our proposed method to construct fuzzy relation based on fuzzy
training data. Actually, this implication method is inspired from
the ink drop spreading concept in active learning method
\cite{shouraki1,murakami} because of three reasons: (i) it has
biological support, (ii) its hardware implementation is very simple
and (iii) it does not obey from exact mathematical techniques.
Figure \ref{figfuzrelcreation} shows this process graphically. In
this figure, fuzzy relation is interpreted as an image where in this
image the intensity of the pixel at location $(x,y)$ is directly
proportional to $\mu_R(x,y)$ and pixels with higher intensity are
depicted darker. When a new training data is being observed, the
value of the fuzzy relation at any point like $(x,y)$ should be
increased (should be made darker). However, increasing amount of
fuzzy relation at point $(x,y)$ should be proportional to membership
grades of observed input and output fuzzy numbers at points $x$ and
$y$ respectively. Since most of naturally created fuzzy numbers have
bell like shapes (gaussian-like membership functions), updating the
fuzzy relation due to this new fuzzy input-output training data can
be equated with the distillation of one ink drop and then spreading
it on the image representing this fuzzy relation. This process is
clearly demonstrated in Fig. \ref{figfuzrelcreation} as well. In a
similar way, updating the fuzzy relation with the next fuzzy
input-output training data will be equal to dropping another ink
drop on the fuzzy relation image. As individual ink drop patterns
overlap, the overlapping area become increasingly darker. The
effective radius of distilling ink drop on the image is to somehow
related to the amount of uncertainty in measuring input and output
numbers. For example, when we can measure input and output data with
infinite precision, they will become simple crisp numbers and
therefore the radius of distilling ink drop should approach zero.

Gradually by the entrance of more and more data (by progressing in
the learning process and spreading more data), a pattern will be
formed on this image.  Finally, this pattern will represent the
fuzzy relation which connects input and output fuzzy variables.

To conclude, any fuzzy inference needs a fuzzy relation which in its
simplest form can be represented by a 2-dimensional matrix (like the
one shown in Fig. \ref{figfuzrelcreation}). Consequently, any fuzzy
inference hardware that one would like to propose should at least
have a structure for storing and manipulating matrixes.

\subsection{Doing inference}
\label{secdoinf} 
\begin{figure}[!t]
\centering \subfigure[]{
\label{figinfaa} 
\includegraphics[width=4in,height=2.5in]{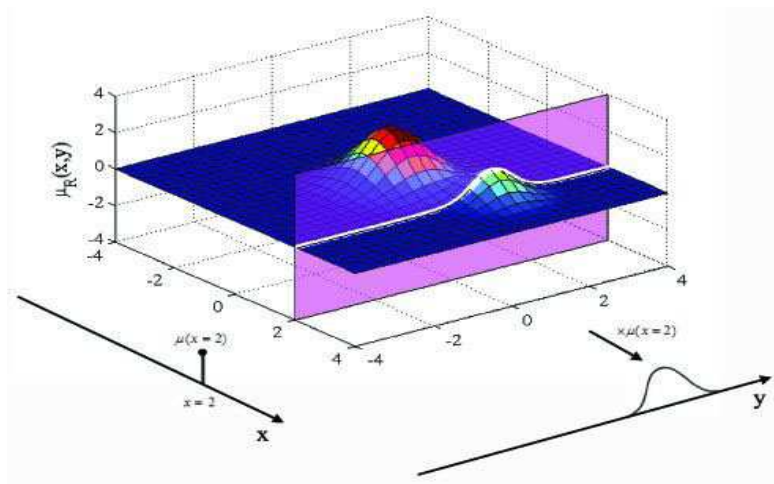}}
\hspace{0.08in} \subfigure[]{
\label{figinfbb} 
\includegraphics[width=5in,height=3in]{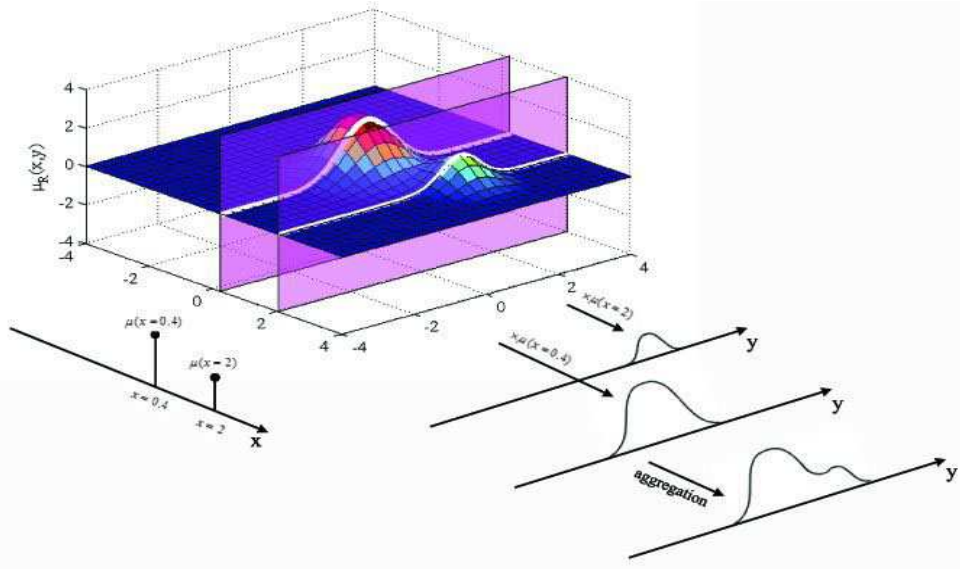}}
 \caption{These figures show how inference is done in our proposed inference system. In our method, each sample of input fuzzy number creates a
distinct output fuzzy number which is the curve of intersection of
the fuzzy relation with the specific plane. Then, these fuzzy
numbers are aggregated (summed) with each other with different gains
to create the output fuzzy number; as the membership grade of one
sample in fuzzy input number be higher, contribution of its
corresponding fuzzy number in the creation of final output fuzzy
number will be more. (a) input of the system is a crisp number. (b)
input of the system is a fuzzy number.}
\label{figinfab} 
\end{figure}

In this section, we explain how inference is done
in our proposed method. For this purpose, assume that fuzzy relation
is created and based on this relation and given input fuzzy number,
we want to determine corresponding output fuzzy number. In
literature, several different approaches have been proposed to
design fuzzy inference systems having linguistic descriptions of
inputs and outputs \cite{mamdani,sugeno}. However, most of these
inference systems have this disadvantage that their hardware
implementation is not simple. For example, consider {\it min} and
{\it max} operators which are frequently in use in fuzzy inference
systems. It is well known that their hardware implementation is not
efficient compared to some other t-norm or t-conorm operators.
Therefore, it is evident that the construction of large scale
systems like human brain based on these inference methods is very
hard if not impossible. Actually, this is because of the fact that
these operators are not inherently consistent with the physical
behavior of circuit elements. To avoid this problem, we propose a
new but unusual inference method as described below. Suppose that we
have a fuzzy relation $R$ and input fuzzy number
$A'=\{(x,\mu_{A'}(x))|x\in \mathbf{X}\}$. Now, we propose to compute
the membership function of output fuzzy number $B'$ corresponds to
the input fuzzy number $A'$ as:
\begin{equation}\label{eqinfmul}
\mu_{B'}(y)=\sum_{x\in\mathbf{X}}\mu_{A'}(x)\times \mu_R(x,y),  \ \
\forall y\in \mathbf{Y}
\end{equation}
where $\mathbf{Y}$ is the universe of discourse of output fuzzy
number $B'$. If we refer back to our proposed structure shown in
Fig. \ref{figa:2} and consider the input of the structure as a
vector of membership grades like
$\mathbf{\mu}_{A'}=[\mu_{A'}(x_1),\mu_{A'}(x_2),\ldots,\mu_{A'}(x_n)]^T$,
in this case output of that structure can be written as:
\begin{equation}\label{eqcomp}
\mathbf{\mu}_{B'}=[\mu_{B'}(y_1),\mu_{B'}(y_2),\ldots,\mu_{B'}(y_m)]^T=\mathbf{\mu_R}\times\mathbf{\mu}_{A'}
\end{equation}
where $T$ is a transposition operator and
$\mathbf{\mu_R}=\{\mu_R(x,y)\}$. Therefore, our proposed inference
method is nothing else than a simple vector to matrix
multiplication. This process is depicted in Fig. \ref{figinfab}
graphically in two cases: (i) input of the system is a crisp number
and (ii) input of the system is a fuzzy number. Note that to have
better understanding, fuzzy relation in this fuzzy system is
considered to be continuous. Figure \ref{figinfaa} shows that when
the input of the fuzzy system is a crisp number, like $x=x_i$ ($x_i$
is equal to 2 in Fig. \ref{figinfaa}), output of the system which is
acquired by our proposed inference method will be the curve
(representing output fuzzy number) of intersection of the fuzzy
relation with the plane $x=x_i$ multiplied by $\mu(x=x_i)$. Figure
\ref{figinfbb} shows the other case in which the input of the system
is a fuzzy number. However, without loos of generality and to better
illustrate the concept behind our inference method, input fuzzy
number is considered to be discrete with only two non-zero samples
at points $x=x_i(=2)$ and $x=x_j(=0.4)$ with membership grades of
$\mu_{A'}(x=x_i)$ and $\mu_{A'}(x=x_j)$ respectively. When this
input is applied to the structure, output fuzzy number $B'$ will be
the weighted sum of two fuzzy numbers obtained by intersecting of
the fuzzy relation with planes $x=x_i$ and $x=x_j$ which can be
written as:
\begin{equation}
B'=\{(y,\mu_{B'}(y))|y\in\mathbf{Y}\}
\end{equation}
where
\begin{equation}\label{eqinfexam}
\mu_{B'}(y)=\mu_{A'}(x_i)\times R(x_i,y)+\mu_{A'}(x_j)\times
R(x_j,y), \ \forall y\in \mathbf{Y}
\end{equation}

This equation means that each sample of input fuzzy number creates a
distinct output fuzzy number which is the curve of intersection of
the fuzzy relation with the specific plane. Then, these fuzzy
numbers are aggregated (summed) with each other with different gains
to create the output fuzzy number; as the membership grade of one
sample in fuzzy input number be higher, contribution of its
corresponding fuzzy number in the creation of final output fuzzy
number will be more. Therefore, our proposed inference method (Eq.
\ref{eqinfmul} or Eq. \ref{eqcomp}) which is only a simple vector to
matrix multiplication is not so much irrelevant and illogical. Note
that since multiplying the left-hand side of Eq. \ref{eqinfexam} by
a constant will not change the output fuzzy number (its
defuzzification before and after this multiplication will result in
a same crisp number), the height of the output fuzzy number does not
need to be equal to 1.

In the following sections, we will show how this proposed inference
method can be simply implemented by using memristor crossbar
structure. In addition, the reason that made us to consider this
system as a neuro-fuzzy one will be explained as well.

\section{Hardware implementation of proposed neuro-fuzzy system}
\label{hardware} In this section, at first physical properties of
the forth circuit
 element {\it i.e.} memristor will be discussed and then
the working procedure of memristor crossbar and its application will
be explained. Finally, we will show that how the fuzzy inference
system that we developed in previous section can be implemented with
this memristor crossbar structure.

\subsection{Memristor} \label{memristor}

After the first experimental realization of the forth fundamental
circuit element {\it i.e.} memristor \cite{williams} whose existence
was previously predicted in 1971 by Leon Chua \cite{Chua},
extraordinary increased researches are in process in variety of
fields like neuroscience, neural networks and artificial
intelligence. It has become clear that this passive element can have
many potential applications such as non-volatile memory construction
\cite{waser}, creation of analog neural network and emulation of
human learning \cite{pershin}, building programmable analog circuits
\cite{Shin, perShin22,farelsevmemarith}, constructing hardware for
soft computing tools \cite{farIEEE}, implementing digital circuits
\cite{kuekes} and in the field of signal processing \cite{Mouttet1,
Mouttet2}.

Memristor, different from other electrical elements namely resistor,
capacitor and inductor, denotes the relationship between
flux($\varphi$) and electric charge ($q$) as \cite{Chua}:
\begin{equation}\label{eq1}
d\varphi=Mdq.
\end{equation}

By rewriting this equation, memristance of the memristor can be
expressed as:
\begin{equation}\label{eq2}
M(q)=\frac{d\varphi/dt}{dq/dt}=\frac{v(t)}{i(t)},
\end{equation}
which shows that the unit of memristance is ohm. In fact, A
memristor can be thought of as a resistive device that its
resistance varies in dependence of its current or flux.

\begin{figure}[!t]
\centering 
{
\includegraphics[width=3in,height=1.3in]{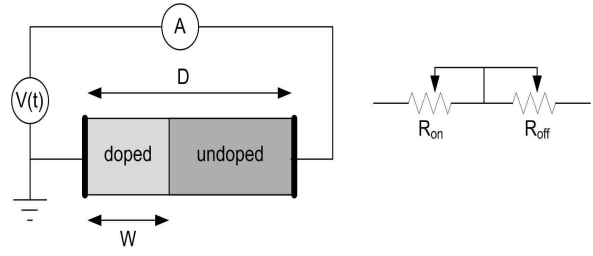}}
\caption{Structure of the memristor reported by HP researchers and
its equivalent circuit model \cite{williams}.}
\label{fig1} 
\end{figure}

Memristor is an electrically switchable semiconductor thin film
sandwiched between two metal contact with a total length of $D$ and
consists of doped and un-doped regions which its physical structure
with its equivalent circuit model is shown in Fig.~\ref{fig1}
\cite{williams}. The internal state variable $w$ determines the
length of doped region with low resistance against un-doped region
with high resistivity. This internal state variable and consequently
the total resistivity of the device can be changed by applying
external voltage bias $v(t)$ \cite{williams2}. If the doped region
extends to the full length $D$, the total resistance of the device
will be at its lowest level denoted as $R_{on}$ and if the un-doped
region extends to the full length $D$, the total resistance of the
device will be at its highest level namely $R_{off}$. For example,
the mathematical model for the total resistance of the memristor
reported by HP can be written as \cite{williams}:
\begin{eqnarray}\label{eq3}
  M(w)&=&R_{on}\frac{w}{D}+R_{off}\left(1-\frac{w}{D}\right),\nonumber \\
  w(t)&=&w_0+\frac{\mu_vR_{on}}{D}q(t),
\end{eqnarray}
where $w_0$ is the initial state for state variable $w$, $\mu_v$ is
the average ion mobility and $q(t)$ is the amount of electric charge
(integral of current) that has passed through the device. Above
equations show that passing current from memristor in one direction
will increase the memristance of the memristor while changing the
direction of the applied current will decrease its memristance. In
addition, it is obvious that in this element, passing current in one
direction for longer period of time (which means $q(t)$ has higher
absolute value) will change the memristance of the memristor more.
Moreover, by setting the passing current into zero, the memristance
of the memristor will not change anymore. Finally, note that
determining the memristance of the memristor at any time can be done
by passing a small current through the memristor and measuring the
dropped voltage across it (see Eq. \ref{eq2}).

As a result, memristor is nothing else than the analog variable
resistor where its resistance can be adjusted by changing the
direction and the duration of the applied voltage. Therefore,
memristor can be used as a storage device in which analog values can
be stored as a memristance instead of voltage or charge. Here it
should be emphasized that memristor-based storage devices have some
advantages compared to those storage devices which use capacitors.
At first, the former can be fabricated much denser than the later
one through the nano-crossbar technology \cite{crossbar}. Second,
memristor can hold the stored data unchanged theoretically for an
infinite period of time without refreshing \cite{memnonvol}.
Finally, unlike capacitors, memristors can be used in the memristor
crossbar structure which has so many potential applications (one of
them is demonstrated in this paper).

\begin{figure}[!t]
\centering 
{
\includegraphics[width=3in,height=2.5in]{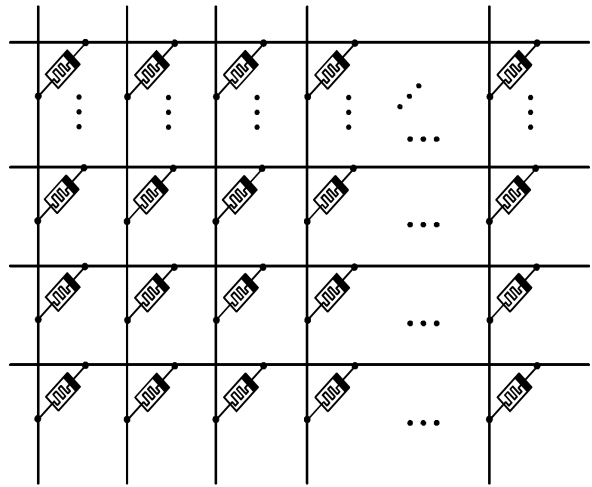}}
\caption{A typical memristor crossbar. In this crossbar, memristor
is formed in each crosspoint.}
\label{fig2} 
\end{figure}

A crossbar array basically consists of two sets of conductive
parallel wires intersecting each other perpendicularly. The region
where a wire in one set crosses over a wire in the other set is
called a crosspoint (or junction). Crosspoints are usually separated
by a thin film material which its properties such as its resistance
can be changed by controlling the voltage applied to it. One of such
materials is memristor which is used in our proposed crossbar-based
circuits in this paper. Figure \ref{fig2} shows a typical memristor
crossbar. In this circuit, memristors which are formed at
crosspoints are depicted explicitly to have better visibility. In
this crossbar, memristance of any memristor can simply be changed by
applying suitable voltages to those wires that memristor is
fabricated between them. For example, consider the memristor located
at coordinate (1, 1) (crossing point of the first horizontal and the
first vertical wires) of the crossbar. Memristance of this memristor
can be decreased by applying a positive voltage to the first
vertical wire while grounding the first horizontal one (or
connecting to negative voltage). Dropping a positive voltage across
the memristor will cause the current to pass through it and
consequently, the memristance of this passive element will be
decreased. In a similar way, memristance of this memristor can be
increased by reversing the polarity of applied voltage. As stated
before, application of higher voltages for longer period of time
will change the memristance of the memristor more. This means that
the memristance of any memristor in the crossbar can be adjusted to
any predetermined value by the application of suitable voltages to
the specific row and column of the crossbar. However, here it should
be emphasized that in this paper, there will be no need to adjust
the memristance of the memristors accurately and only increasing or
decreasing the memristance of the memristor will be sufficient.

To summarize, memristor crossbar is a 2-dimensional grid that analog
values can be stored in its crosspoints through the memristance of
the memristors. Consequently, it seems that the memristor crossbar
is a perfect structure to construct and store 2-dimensional patterns
like fuzzy relations. In this case, vertical and horizontal wires of
the crossbar will represent different discrete values of input and
output variables respectively and memristors in crosspoints will
have the role of entries of 2-dimensional matrixes representing
fuzzy relations.

Against these mentioned advantages, memristor crossbar structure has
some drawbacks. First of all, reading the memristance of the
memristors at any point of the crossbar is partly difficult. But the
second and the most important disadvantage of memristor crossbar
structure relates to the inherent physical limitation of memristor.
As stated before, memristance of the memristor can be between
$R_{0ff}$ and $R_{on}\neq0$. Therefore, it seems that storing analog
values through the memristance of the memristors needs a linear
mapping from input space to interval $[R_{on}, R_{off}]$. In the
next section, we show how these problems can be solved by slightly
modifying the structure of Fig. \ref{fig2} and how this structure
can be used to construct and store fuzzy relations.

\subsection{Using memristor crossbar to do fuzzy inference}

\label{basic} 
\begin{figure}[!t]
\centering 
{
\includegraphics[width=6in,height=4in]{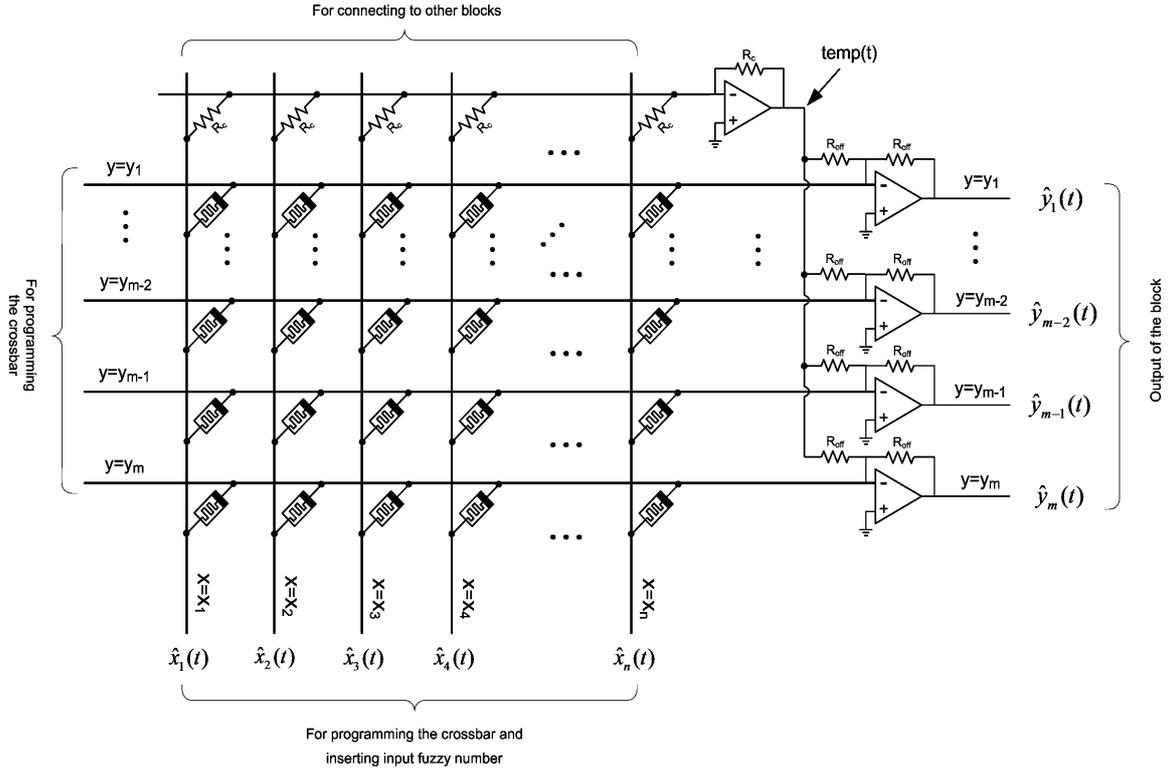}}
\caption{Memristor crossbar-based analog circuit which is proposed
to implement the fuzzy inference method developed in this paper.
Each vertical and horizontal wire of the crossbar represents
distinct value of input and output variables respectively. In this
structure, fuzzy relation between input and output variables be
formed in the memristors of the crossbar.}
\label{fig3} 
\end{figure}

Figure \ref{fig3} shows the memristor crossbar-based
analog circuit that we have proposed to construct fuzzy inference
system introduced in previous section. As can be seen later, this
structure has biological support and can be considered as a
single-layer fuzzy neural network which conforms with the Hebbian
learning rule but with very new interesting properties compared to
today's conventional neural networks.

Figure \ref{fig3} consists of a simple crossbar that each of its
rows is connected to the negative input terminal of a simple opamp.
Through these opamp-based circuits added to rows of the crossbar we
will be able to read the pattern stored in the crosspoints of the
crossbar. Since these opamps have a fixed resistor with resistance
$R_{off}$ (maximum memristance value of the memristors used in the
crossbar) as a feedback, combination of each of these opamps and
those memristors which are connected directly to the same horizontal
wire that opamp is connected to it creates a simple opamp-based
summing circuit. Note that in the uppermost row of the crossbar,
instead of memristor, a resistor with resistance $R_c$ is fabricated
in each crosspoint. Since this row of the crossbar is connected to
an opamp with a feedback resistor $R_c$, it again forms a simple
inverting summing circuit that adds inputs of the crossbar with the
same gain {\it i.e.} -1. Therefore, at any time output of this
summing circuit, {\it i.e.} $temp(t)$, can be written as:
\begin{eqnarray}\label{eq4}
temp(t)=-\sum_{j=1}^n \hat{x}_j(t),
\end{eqnarray}
where $\hat{x}_j(t)$ is the input signal (represented by voltage)
connected to the {\it j}th vertical wire of the crossbar and $n$ is
the number of columns of the crossbar. Note that the hat sign in Eq.
\ref{eq4} is used to distinguish inputs and outputs of the structure
from those concepts which are assigned to rows and columns of the
circuit. The reason of using this summing circuit will become clear
in the rest of this section. Now, by applying standard opamp circuit
analysis techniques, output voltage of the opamp connected to the
{\it i}th row of the crossbar can be expressed as:
\begin{eqnarray}\label{eq5}
\hat{y}_i(t)=-\left(\sum_{j=1}^n
\hat{x}_j\frac{R_{off}}{M_{ij}(t)}+temp(t)\right) \text{for} \ i=1,
2, \ldots, m,
\end{eqnarray}
where $M_{ij}(t)$ is the current memristance of the memristor at
coordinate $(i,j)$ of memristor crossbar and $m$ is the number of
rows of the crossbar. Note that to write this equation, memristor is
treated like a simple resistor. This is true only when the
memristance of the memristor be almost constant during the
calculation of output voltages. This can be guaranteed by applying
input voltages $\hat{x}_j$ (for $j=1, 2, \ldots, n$) to columns of
the crossbar for very short period of time during the execution
phase. Another condition that this structure should satisfy is that
the memristance of all memristors in the crosspoints of the crossbar
should initially be equal to $R_{off}$. In fact, in all of our
proposed structures in this paper, storing any value at coordinate
$(i,j)$ of the crossbar is done by decreasing the initial
memristance of the memristor located at this coordinate by the given
value. In other words, current stored value at coordinate $(i,j)$
denoted by $\Delta M_{ij}(t)$ can be written as:
\begin{eqnarray}\label{eq6}
\Delta M_{ij}(t)=R_{off}-M_{ij}(t),\qquad \forall i, j: 1\leq j\leq
n \ \text{and}\  1\leq i\leq m,
\end{eqnarray}

By this trick, there will be no need to use linear mapping as
mentioned in previous section. By substituting \ref{eq6} and
\ref{eq4} into \ref{eq5} we will get:
\begin{eqnarray}\label{eq6-1}
\hat{y}_i(t)=-\left(\sum_{j=1}^n
\hat{x}_j\frac{R_{off}}{R_{off}-\Delta M_{ij}(t)}-\sum_{j=1}^n
\hat{x}_j(t)\right) \text{for} \ i=1, 2, \ldots, m,
\end{eqnarray}

Now, by assuming that the stored value in every crosspoint is always
much smaller than $R_{off}$, {\it i.e.} $\frac{\Delta
M_{ij}(t)}{R_{off}}\ll 1\ (\forall i,j)$, which can be satisfied by
scaling data before storing, Eq. \ref{eq6-1} can be simplified to:
\begin{eqnarray}\label{eq6-2}
\hat{y}_i(t)&=&-\left(\sum_{j=1}^n \hat{x}_j\frac{1}{1-\frac{\Delta
M_{ij}(t)}{R_{off}}}-\sum_{j=1}^n \hat{x}_j(t)\right)\simeq
-\left(\sum_{j=1}^n \hat{x}_j\left(1+\frac{\Delta
M_{ij}(t)}{R_{off}}\right)-\sum_{j=1}^n
\hat{x}_j(t)\right)\nonumber\\&=&-\sum_{j=1}^n \hat{x}_j\frac{\Delta
M_{ij}(t)}{R_{off}}\qquad \text{for} \quad i=1, 2, \ldots, m,
\end{eqnarray}
or equivalently can be expressed in matrix form as:
\begin{eqnarray}\label{eq6-3}
\hat{\mathbf{y}}(t)=\alpha\Delta\mathbf{M}(t)\left(\hat{\mathbf{x}}(t)\right)^T,
\end{eqnarray}
where $\alpha=\frac{-1}{R_{off}}$,
$\hat{\mathbf{x}}(t)=\left[\hat{x}_1(t), \hat{x}_2(t), \ldots,
\hat{x}_n(t)\right]$, $\hat{\mathbf{y}}(t)=\left[\hat{y}_1(t),
\hat{y}_2(t), \ldots, \hat{y}_m(t)\right]^T$, and $\Delta
\mathbf{M}(t)=\{\Delta M_{ij}(t)\}$. It is evident that although we
will use this structure for performing fuzzy operations, any matrix
multiplication in the form of Eq. \ref{eq6-3} can be done by this
structure.

%
By comparing Eqs. \ref{eq6-3} and \ref{eqcomp} (or equivalently by
comparing structures of Fig. \ref{fig3} and Fig. \ref{figa:2}) it
becomes clear that if somehow we can store fuzzy relation in the
crosspoints of the crossbar (as matrix $\Delta\mathbf{M}(t)$) our
proposed structure will be a perfect analog circuit to perform our
proposed inference method (note that in Eq. \ref{eqcomp},
$\mathbf{\mu}_{A'}$ was a column vector while in Eq. \ref{eq6-3}
$\hat{\mathbf{x}}(t)$ is a row vector). For this purpose, it is
sufficient to consider that the {\it i}th column and the {\it j}th
row of the crossbar represent concepts $x=x_i$ and $y=y_j$
respectively similar to inputs and outputs of the system shown in
Fig. \ref{figa:2} and then connect the membership grades of newly
observed input fuzzy number to the inputs of this structure. In this
case, the membership grades of the corresponding output fuzzy number
will emerge at outputs of the structure instantaneously. As another
example, consider the case in which the row vector $\mathbf{v}_k$
with only one non-zero element at position $k$ is being applied as
an input to the structure where fuzzy relation $\Delta\mathbf{M}(t)$
is programmed in its crosspoints. In this case, the output of the
structure will be $\alpha
\left(v_k\times\Delta\mathbf{m}_k(t)\right)$ where
$\Delta\mathbf{m}_k(t)$ is the {\it k}th column of matrix
$\Delta\mathbf{M}(t)$ and $v_k$ is the value of the {\it k}th entry
of vector $\mathbf{v}_k$. Note that this example is equivalent to
the inferential process depicted in Fig. \ref{figinfaa}. As stated
in that section, multiplication of output fuzzy number
($\Delta\mathbf{m}_k(t)$ which is the intersection of stored
2-dimensional fuzzy relation ($\Delta\mathbf{M}(t)$) with the plane
$x=v_k$) with a constant, {\it i.e.} $\alpha$, will not degrade the
generated output fuzzy number significantly or it can be easily
compensated by using some auxiliary circuits.

As stated before, domain and resolution of input and output
variables can be increased by adding more horizontal and vertical
wires to the structure. Simplicity of the circuit and computing in
realtime are some of the advantages of our proposed hardware.

To summarize, proposed hardware of Fig. \ref{fig3} can do vector to
matrix multiplication and therefore can do any of the inferential
processes previously described in Sec. \ref{secdoinf}. The only
question that we should answer is how to program the memristors in
the crossbar or how to create and store fuzzy relation in this
memristor crossbar-based structure. This will be addressed in the
next subsection.

\subsection{creating and storing fuzzy relations on the proposed
memristor crossbar-based structure}

In this subsection, we will consider two cases to construct a fuzzy
relation on the proposed structure of Fig. \ref{fig3}: (i) input and
output fuzzy sets are available and (ii) instead of input and output
fuzzy sets some input-output training data are in hand.
\subsubsection{creating binary fuzzy relation based on given input and output fuzzy sets}

In the first case, assume that we have input fuzzy set $A$ and
output fuzzy set $B$ with membership functions $\mu_A(x)$ and
$\mu_B(y)$ respectively and we want to construct a fuzzy relation
between these two fuzzy sets based on fuzzy implication of Eq.
\ref{eqfr}. To create a fuzzy relation from $\mu_A(x)$ and
$\mu_B(y)$ in the memristor crossbar structure of Fig. \ref{fig3},
it is sufficient to interpret membership degrees in each fuzzy set
as a voltage and apply them to their corresponding rows and columns
of the memristor crossbar for $t_0$ second(s). To be exact, each
element of $\mu_A(x)$ should be connected to its corresponding
column in the crossbar and the negative of each element of
$\mu_B(y)$ should be connected to its corresponding row in the
crossbar (remember that as depicted in Fig. \ref{figa:2}, each input
and output terminal of the structure corresponds to exclusive value
or concept of input and output fuzzy variables respectively). For
example, if sets $A$ and $B$ be defined as:
\begin{eqnarray}\label{eqdefab}
A=\left\{(x_1,\mu_A(x_1)), (x_2,\mu_A(x_2)), \ldots, (x_n,\mu_A(x_n))\right\}\\
B=\left\{(y_1,\mu_B(y_1)), (y_2,\mu_B(y_2)), \ldots,
(y_m,\mu_B(y_m))\right\}
\end{eqnarray}
then to create a fuzzy relation which is connecting these input and
output fuzzy sets on this structure, these following tasks should be
done:

\begin{itemize}
\item interpret membership grades as a voltage signal and connect $\mu_A(x_1)$ to the vertical wire of the crossbar
representing concept $x=x_1$, $\mu_A(x_2)$ to the vertical wire of
the crossbar representing concept $x=x_2$ and so on.
\item at the same time, connect $-\mu_B(y_1)$ to the horizontal wire of the crossbar
representing concept $y=y_1$, $-\mu_B(y_2)$ to the horizontal wire
of the crossbar representing concept $y=y_2$ and so on.
\item only wait for $t_0$ second(s) and then remove applied voltages
from the crossbar.
\end{itemize}

\begin{figure}[!t]
\centering 
{
\includegraphics[width=5.8in,height=3.1in]{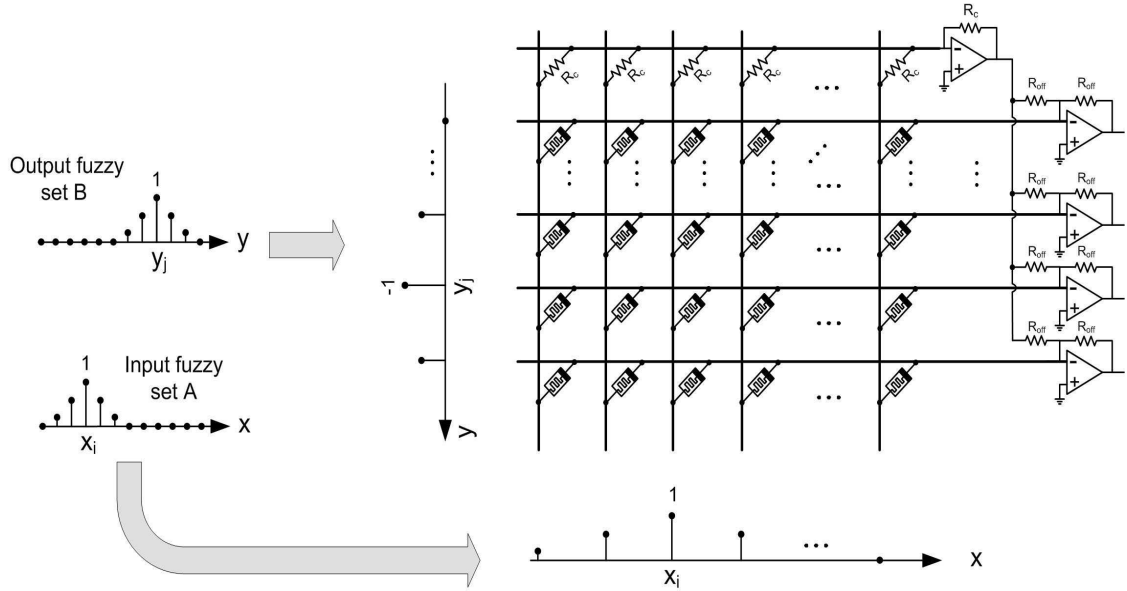}}
\caption{This figure shows the procedure of the creation of the
fuzzy relation on the memristor crossbar of the circuit of Fig.
\ref{fig3}. For this purpose it is only sufficient to connect
samples of input and output fuzzy sets to their corresponding
vertical or horizontal wires for $t_0$ second(s).}
\label{figfuzrela} 
\end{figure}

This process is depicted in Fig. \ref{figfuzrela} for two typical
fuzzy sets. Simultaneous application of positive and negative
voltages to rows and columns of the crossbar respectively will cause
the current to pass through the memristors in crosspoints. Amount of
current passing through the memristor located at the intersection
point of {\it j}th column and {\it i}th row of the crossbar
corresponding to concepts $x=x_j$ and $y=y_i$ respectively (or
equivalently through the memristor at coordinate $(i,j)$ of the
crossbar) is directly proportional to the dropped voltage over this
passive element or equivalently to
$\left(\mu_A(x=x_j)+\mu_B(y=y_i)\right)$. As this term increases,
memristance of the memristor at coordinate $(i, j)$ will decrease
more during this $t_0$ second(s). Assuming the HP model for the
memristors of the crossbar, memristance of this memristor after this
$t_0$ second(s) can be written as \cite{memrismodel}:
\begin{eqnarray}\label{eq8-11}
M_{ij}(t)|_{t=t_0}=\sqrt{R_{off}^2-\beta
(\mu_A(x=x_j)+\mu_B(y=y_i))t_0}, \ \forall i, j: 1\leq i\leq m \
\text{and}\ 1\leq j\leq n,
\end{eqnarray}
where $\beta$ is a constant defined as:
\begin{eqnarray}\label{eq8-112}
\beta=\frac{2\mu_vR_{on}(R_{off}-R_{on})}{D^2}
\end{eqnarray}

Note that to obtain Eq. \ref{eq8-11}, initial memristance of the
memristors, $M_{ij}(t=0)$, as expected is assumed to be $R_{off}$.
Using Eq. \ref{eq8-11} and Eq. \ref{eq6}, amount of change of
$M_{ij}(t)$ during this $t_0$ second(s) which is equal to the stored
value at coordinate $(i,j)$ of the crossbar will become:
\begin{eqnarray}\label{eq8-12}
&&\Delta M_{ij}(t)|_{t=t_0}
=\mu_R(x=x_j,y=y_i)=M_{ij}(t)|_{t=0}-M_{ij}(t)|_{t=t_0}
=R_{off}-M_{ij}(t)|_{t=t_0}\nonumber\\&=&R_{off}-\left(\sqrt{R_{off}^2-\beta
(\mu_A(x=x_j)+\mu_B(y=y_i))t_0}\right)\ \forall i, j: 1\leq i\leq m\
\text{and}\  1\leq j\leq n,
\end{eqnarray}

Comparing Eqs. \ref{eq8-12} and \ref{eqfr} shows that the
implication function $f(\cdot)$ considered in Eq. \ref{eqfr} for
this specific memristor crossbar structure is:
\begin{eqnarray}\label{eq8-122}
f(\nu)=R_{off}-\sqrt{R_{off}^2-\beta t_0\nu}
\end{eqnarray}

\begin{figure}[!t]
\centering 
{
\includegraphics[width=4in,height=2.2in]{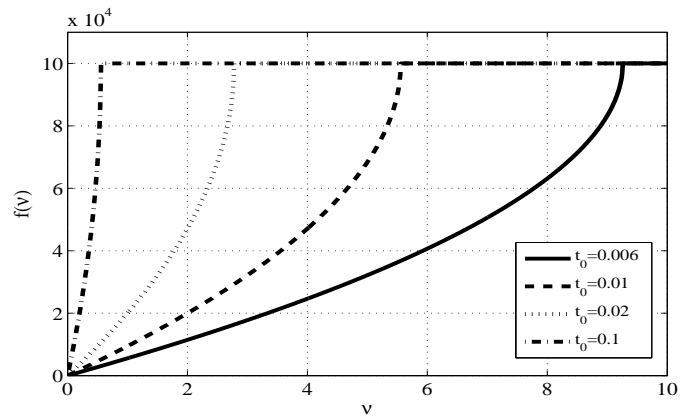}}
\caption{A plot of function $f(\cdot)$ defined in Eq. \ref{eq8-122}
for different values of $t_0$ which shows that this is a
monolithically increasing function.}
\label{figfunfimp} 
\end{figure}

Figure \ref{figfunfimp} shows a plot of function $f(\cdot)$ for
different values of $t_0$ which shows that as expected it is a
monolithically increasing function. For plotting this figure, values
of other parameters are set as following:
$\mu_v=10^{-14}m^2s^{-1}V^{-1}$, $D=10^{-8}$m, $R_{on}=1K\Omega$ and
$R_{off}=100K\Omega$. Although here we tried to obtain mathematical
expression for the implication method of our proposed hardware, it
is clear that while working with this structure, there is no need to
be involved with these exact mathematics. The only thing that we
should do is to connect voltages to rows and columns of the crossbar
and wait for $t_0$ second(s) and then fuzzy relation will
automatically be created and stored on the crossbar. Finally note
that to satisfy the mentioned condition for properly working of the
memristor crossbar structure, {\it i.e.} $\Delta M_{ij}(t)\ll
R_{off},\ (\forall i,j)$, it is necessary to choose a small value
for $t_0$ (for example, in the above test, $t_0=0.0001$ will be a
good choice).

\subsubsection{creating 2-dimensional fuzzy relation based on available training data}
\label{fuzreltrdat} It is well known that one of the major
applications of memristor is in the construction of non-volatile
memories. This is because of the fact that memristance of the
memristor will remain fixed theoretically for an infinite period of
time when there is no applied voltage or current. This can be seen
in Eq. \ref{eq3} by setting $q(t)$ to zero. This means that the
created fuzzy relation on the crossbar will remain unchanged without
refreshing (which is required for most of currently available
capacitor-based memories). This property of memristor says that
final memristance of the memristor after the application of specific
voltage (or current) will act as a initial memristance of the
memristor during the application of next voltage (or current). This
means that effects of these sequentially applied voltages on the
memristance of the memristor will be added to each other. By this
explanation, it would be logical to think that one simple way for
the construction of fuzzy relation on the proposed hardware based on
available input-output fuzzy training data can be the repetition of
the process described in previous subsection for each of these
training data. To illustrate this process a bit more, consider that
$N$ input-output fuzzy training data like $\{A'_i, B'_i\}$ for
$i=1,2,\ldots,N$ are available and we want to construct a pattern on
our memristor crossbar circuit corresponding to the fuzzy relation
between input and output fuzzy variables based on these training
data. For this purpose, it is sufficient to behave as follows:

For $i=1,2,\ldots,N$ repeat these steps:
\begin{itemize}
\item connect membership grades of fuzzy number $A'_i$ and the negative of membership grades of fuzzy number $B'_i$ to
their corresponding columns and rows of the
crossbar respectively.
\item wait for $t_0$.
\end{itemize}

presentation of each of these training data will decrease the
memristance of the memristors in some areas of the crossbar.
Therefore, the effect of observing any new training data will be
added to the currently stored pattern in the crossbar. In other
words, entrance of any new training data simply updates the stored
pattern. Hence, by presenting more training data to the structure,
gradually a pattern (fuzzy relation) will begin to form on the
crossbar.

Here, three important aspects of this described procedure should be
emphasized. First, $t_0$ should be decreased by the increase of $N$
to satisfy the condition $\Delta M_{ij}(t)\ll R_{off} \ (\forall
i,j)$. Second, since a crisp number is a special case of fuzzy
number, it is evident that the described procedure will be also
applicable to crisp training data. Creating fuzzy relation based on
crisp data is possible either by using large number of training data
to form a pattern or by converting crisp training data to their
corresponding fuzzy numbers before applying them to the structure
(for example by using gaussian membership function). The third and
the most important note pertains to how to create voltages
corresponding to membership grades of input and output fuzzy numbers
and how to apply(remove) them to(from) inputs and outputs of the
structure. As will be demonstrated in the section of simulation
results as well, there is no need to have any other auxiliary
circuit to perform these tasks. This is because of the fact that our
proposed circuit has this capability that it can be directly
connected to other similar circuits. In this case, output of one
circuit will be the input of the next structure and by this way,
signals will propagate easily in the entire system. At the same
time, based on the current values of input and output of each of
these proposed memristor crossbar-based circuits in the whole
system, stored pattern in their crossbar will automatically be
updated.

\section{why our proposed structure is a neuro-fuzzy system}
\label{reason}

The neuromorphic paradigm is attractive for nanoscale computation
because of its massive parallelism, potential scalability, and
inherent defect- and fault-tolerance \cite{weilu,Chabi}. In
biological systems, the synaptic weights between neurons can be
precisely adjusted by the ionic flow through them and it is widely
believed that the adaptation of synaptic weights enables biological
systems to learn and function. However, before the first physical
realization of memristor, experimental construction of these
neuromorphic systems which consist of neurons and synapses,
especially in the electronic domain, has remained somewhat
difficult. The primary problem was the lack of a small and efficient
circuit that can emulate essential properties of synapses namely:
having low power consumption, ability to be fabricated in high
density (human brain has about $10^{14}$ synapses) and plasticity.

\begin{figure}[!t]
\centering \subfigure[]{
\label{figconvneur:a} 
\includegraphics[width=2.4in,height=2in]{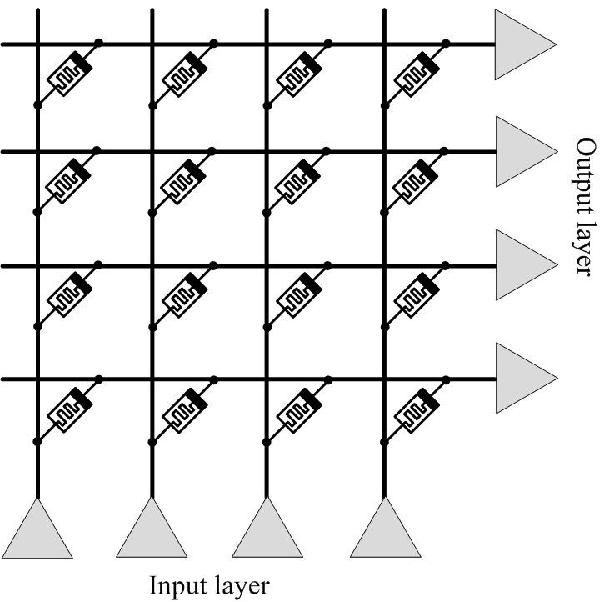}}
\vspace{0.14in} \subfigure[]{
\label{figconvneur:b} 
\includegraphics[width=2.4in,height=2in]{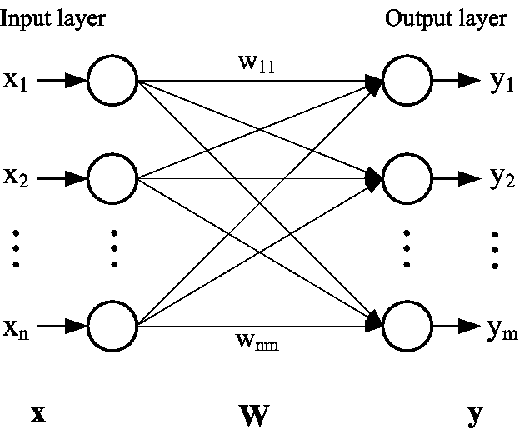}}
 \caption{(a) Memristor crossbar-based structure recently have been proposed for the hardware
 implementation of the third generation of artificial neural networks. (b) Conventional single
 layer neural network consists of neurons with identity activation
 function.}
\label{figconvneur} 
\end{figure}

After the first experimental realization of memristor
\cite{williams}, it became widely accepted that memristor is a good
candidate to emulate synapse. This is because of the fact that
memristors can be fabricated in high density through the crossbar
technology. In addition, similar to biological synapse, the
conductance of the memristor can be changed by passing current from
it or applying voltage to it \cite{weilu,Cantley}. Based of these
evidences, several authors have tried to use memristor as a synapse
and showed that this passive element can facilitate hardware
implementation of artificial neural networks and their corresponding
learning rules such as Spike Time Dependent Plasticity (STDP)
\cite{afifi,Carrasco,Snider}. Figure \ref{figconvneur:a} shows the
memristor crossbar-based structure that has been proposed in almost
all of these works. In this figure, each fabricated memristor at
each crosspoint represents one synapse which connects one
presynaptic to one postsynaptic neuron. In this case, every neuron
(shown by a triangle) in the input layer is directly connected to
every neuron in output layer with unique synaptic weights. If in
Fig. \ref{figconvneur:a}, we denote output voltages of presynaptic
neurons by a row vector $\hat{\mathbf{x}}$, output voltages at
postsynaptic neurons by a column vector $\hat{\mathbf{y}}$ and
weights of synapses by matrix $\mathbf{W}$, the structure of Fig.
\ref{figconvneur:a} implements the following mathematical operation:
\begin{eqnarray}\label{eqneuro1}
\hat{\mathbf{y}}=\mathbf{W}\hat{\mathbf{x}}
\end{eqnarray}
where to obtain this equation, we have assumed identity activation
function for neurons. Note that Eq. \ref{eqneuro1} is a very
familiar equation we have in conventional neural networks like the
typical one depicted in Fig. \ref{figconvneur:b} \cite{faset}.
Therefore, Fig. \ref{figconvneur:a} does not present anything new
but only shows how memristor crossbar can be used as an electrical
representation of synapses weights. Learning can be accomplished in
this structure based on learning methods mostly inspired from
Hebbian learning rule (or equivalently STDP in spiky neural
networks) which in its simplest form says: {\it neurons that fire
together, wire together} \cite{Hebb}. This means that when one input
and one output neuron fire simultaneously, the weight of that
synapse which is connecting these neurons should be increased
(strengthened). For this purpose, in Fig. \ref{figconvneur:a}
memristance of the memristor connecting these neurons should be
modified (actually it should be decreased). However, unfortunately
it is well known that Hebbian learning rule (although is very
simple) is not enough to train neural networks and other complex
methods like backpropagation learning algorithm are needed
\cite{faset}. Moreover, hardware implementation of currently in use
learning methods like STDP on this structure is difficult and needs
some auxiliary circuits \cite{farmemproblem}. In addition, in these
kinds of learning rules synaptic weights may be either increased or
decreased. These mentioned drawbacks have caused the efficient
hardware implementation of neural networks very difficult.

Now, compare the structure of Fig. \ref{figconvneur:a} and Eq.
\ref{eqneuro1} with Fig. \ref{fig3} and Eq. \ref{eq6-3}
respectively. In addition to their structural similarities, both of
these structures perform the same task (vector to matrix
multiplication) during the execution phase. However, in learning
phase, we can see some differences. In the structure representing
hardware implementation of artificial neural networks such as the
one depicted in Fig. \ref{figconvneur:a}, learning is usually
carried out based on methods like back propagation or STDP. On the
other hand, learning in our proposed method is based on the creation
of fuzzy relation between input and output variables or concepts.
However, here we will show that \emph{learning based on the creation
of fuzzy relation is the same as fuzzy learning in artificial neural
networks but}. To illustrate this theorem better, consider the
creation of fuzzy relation on the circuit of Fig. \ref{figa} based
on available fuzzy training data described in Section
\ref{fuzreltrdat}. In this process, simultaneous application of
input and output fuzzy numbers to inputs and outputs of the
structure is equivalent with the simultaneous firing of input and
output neurons and the value of applied membership grades to each of
input or output terminals actually determines the firing strength of
each neuron. By this statement we mean that each simultaneous firing
of input and output neurons generates one fuzzy input-output
training data. Therefore, in our belief, all of neurons are active
at the same time during learning phase but with different confidence
degrees where output of each neuron determines the confidence degree
of its activation. In fact, when we connect $\mu(x=x_i)$ to the
column of the crossbar representing concept $x=x_i$ it means that
the neuron in input layer which represents concept $x=x_i$ is firing
with the confidence degree of $\mu(x=x_i)$. By interpreting applied
membership grades as a firing strength of neurons, the concept of
ink drop spread introduced in Section \ref{inkdrop} to create fuzzy
relation on the crossbar (see Fig. \ref{figfuzrelcreation}) becomes
equal to primary Hebbian learning rule in neural networks. For
example, Fig. \ref{figfuzrelcreation} from another point of view
says that weights of those connections which are connecting
simultaneously firing input and output neurons should be
strengthened. However, the amount of increase of the weight of each
connection should be proportional to the firing strength of those
input and output neurons. When confidence degrees of firing of both
input and output neurons are high, the weight of the connection
connecting these input and output neurons should be increased more
(see Eq. \ref{eqfr} or Eq. \ref{updateink}). Note that if we have
crisp input-output training data instead of fuzzy one, ink drop
spread method in Fig. \ref{figfuzrelcreation} will become completely
the same as Hebbian learning rule in conventional neural networks.
Therefore, it seems that our proposed fuzzy inference system
implements the same function as neural networks and also learns in a
similar way. In this case, a simple question may arise: {\it what is
the benefit of our proposed structure and inference method compared
to conventional neural networks?}. To answer this question, the
reader should note that in reality, by these explanations we tried
to show that neural networks and fuzzy are not two disjoint fields.
In fact, we strongly believe that the reason which causes Hebbian
learning not to be able to learn neural networks properly is the
misinterpretation of the nature of input and output signals in
neural networks.

Let's look at neural networks from another point of view for a short
time. Suppose that each neuron represents one and only one concept
like $x=x_i$ where $x$ can be a numerical or linguistic variable.
Note that in this situation, spatial location of each neuron in the
entire system determines its corresponding concept. In addition,
assume that output (firing strength) of each neuron in the network
specifies the confidence degree of the activation of that neuron or
its corresponding concept. In this case, combination of those
neurons (or concepts) lied in one layer and their corresponding
confidence degrees (outputs) simply creates a fuzzy number. Now,
application of primary Hebbian learning method to this described
neural network with this kind of input and output signals will
result in a creation of fuzzy relation on the matrix representing
connection weights. Note that changing the functionality of neural
networks to work with confidence or membership degrees is not in
contrast with biological findings but in return it offers some
interesting properties. First of all, as we will illustrate in the
section of simulation result, in spite of conventional neural
networks a simple primary Hebbian learning method (or equivalently
creating fuzzy relation) is completely enough to learn these kinds
of networks. Consequently, as we showed in this paper their hardware
implementation becomes more simple than other learning methods in
artificial neural networks such as STDP or backpropagation. Second,
since in Hebbian learning the connection weights are only
strengthened, they will always be non-negative. Note that we have
the same case in our proposed fuzzy inference system since the
created fuzzy relation on the crossbar is always non-negative. It is
clear that working with non-negative weights is much simpler than
working with weights which can have both positive and negative
values. Third, if after the training of the conventional neural
networks, we plot the matrix which is holding connection weights as
a surface it will have no meaningful shape and no information can be
obtained from it. However, corresponding surface in a neural network
which works with confidence or membership degrees is a fuzzy
relation where its shape describes the overall behavior of input and
output variables versus each other. Remember that human brain
remembers most concepts and relations through the images (surfaces).
Fourth, in the field of artificial neural networks and based on
biological findings, it is common to put a threshold on the outputs
of neurons. However, putting threshold on confidence or membership
degrees (output of our proposed structure) seems to be more logical
than putting threshold on meaningless output values in conventional
neural networks. Finally, it is interesting to note that that in
Sec. \ref{fuzreltrdat} we told that by increasing the number of
training data, $t_0$ should approach zero. In this case, signals
that propagate in the network will become similar to spike and
therefore we would have spiking neural networks.

To summarize, in this section we tried to show that it is possible
to look at the working procedure of  neural networks from another
aspect. If we accept that biological neurons transmit information in
the form of confidence degrees, then the computational task which is
done in conventional neural network will become equal to our
proposed fuzzy inference method with extra advantages.

\section{simulation results}
\label{simulation} 
\begin{figure}[!t]
\centering \subfigure[]{
\label{figfunfafb:a} 
\includegraphics[width=3.2in,height=2.4in]{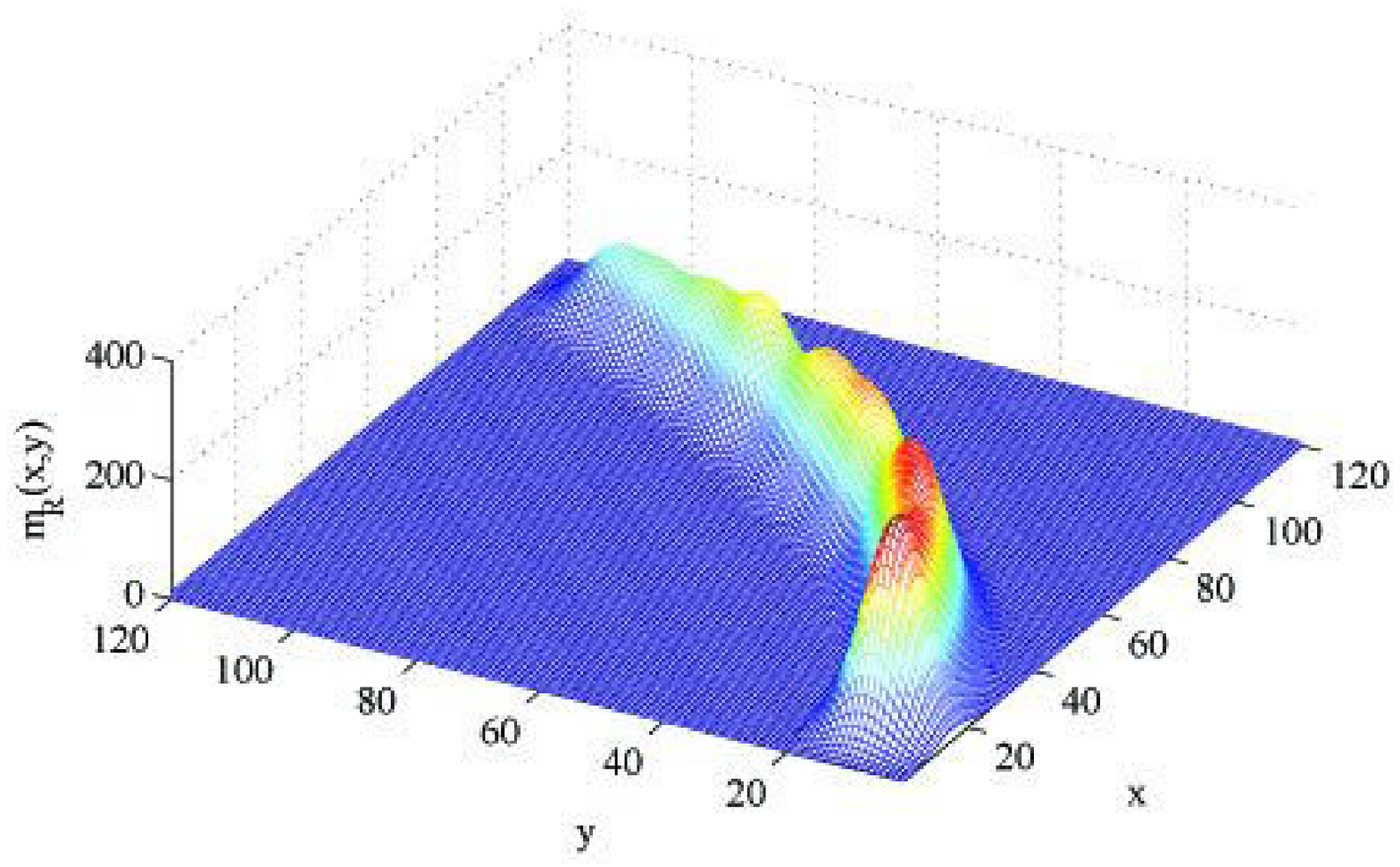}}
\vspace{0.14in} \subfigure[]{
\label{figfunfafb:b} 
\includegraphics[width=3.2in,height=2.4in]{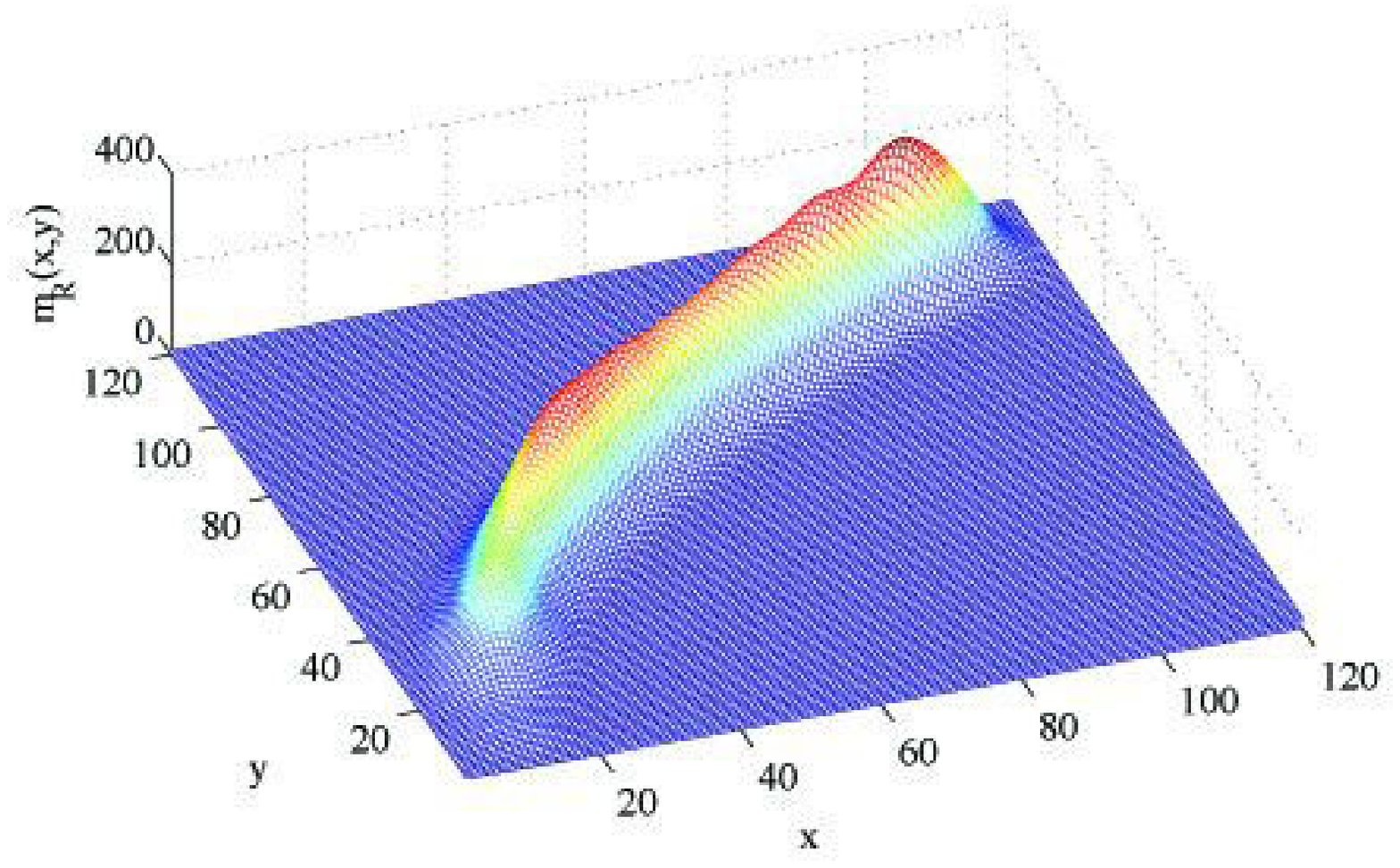}}
 \caption{Constructed fuzzy relations on the crossbar of the circuit of Fig. \ref{fig3} based on
 artificially created fuzzy input-output training data. (a) Fuzzy relation of function $y=f_1(x)=x^2$.
 (b) Fuzzy relation of function $y=f_2(x)=\sqrt{x}$.}
\label{figfunfafb} 
\end{figure}

In this section, we want to verify the efficiency
of our proposed hardware and fuzzy inference method by conducting
several simulations. In the first simulation, we show how
mathematical functions can be constructed in an unprecise manner and
how function composition can be done in our proposed structure. For
this purpose, consider the creation of two fuzzy relations
representing two different functions $y=f_1(x)=x^2$ and
$y=f_2(x)=\sqrt{x}$ on two distinct samples of the circuit shown in
Fig. \ref{fig3}. To construct each of these functions on the circuit
of Fig. \ref{fig3} near 500 input-output fuzzy training data are
used. These training data are created artificially by firstly
generating 500
 crisp input-output training data uniformly distributed on input
 domain and then converting them to their corresponding fuzzy
 numbers by using gaussian membership function. Although in this paper we have used
 gaussian membership function for the fuzzification of crisp training data, repeating
 this test showed that any other membership function can also be used without degrading output results significantly.
 Figures \ref{figfunfafb:a} and
 \ref{figfunfafb:b}
 show the constructed fuzzy relations on the memristor crossbar of the circuit of
 Fig. \ref{fig3} based on these training data by following the procedure described in Section \ref{fuzreltrdat}.
 Note that these figures indicate final
 stored values (or created fuzzy relations) on the crossbars, {\it i.e.} matrix $\Delta \mathbf{M}$, after the accomplishment of training
 process where they are plotted as a continues surface to have better visibility. In this test, it is assumed that
 input variable of both functions $f_1$ and $f_2$, {\it i.e.} $x$, is bounded between 0 and 1 and memristor crossbar of
 the circuit of Fig. \ref{fig3} has 100 rows and
 100 columns (100 vertical and 100 horizontal wires). Other
 parameters are set as follows: $\mu_v=10^{-14}m^2s^{-1}V^{-1}$, $D=10^{-8}$m, $R_{on}=1K\Omega$, $t_0=0.0001$ and
$R_{off}=100K\Omega$. These crossbars are simulated in HSPICE
software by utilizing the SPICE model proposed in \cite{Biolek} for
memristors. Figure \ref{figfunfafb:a} and
 \ref{figfunfafb:b} demonstrate that
 application of Hebbian learning to our proposed structure which we showed that it is equal to the creation of
 fuzzy relation creates a
 meaningful surface on the memristor crossbar since
 it is easy to recognize shapes of functions $f_1(x)=x^2$ and $f_2(x)=\sqrt{x}$
  in these figures. Therefore, it is clear that in
  our structure, relations between input and output variables are stored in the
  system based on their shapes and not through their exact
  mathematical formulas.

\begin{figure}[!t]
\centering \subfigure[]{
\label{figcomb:a} 
\includegraphics[width=5.3in,height=3.1in]{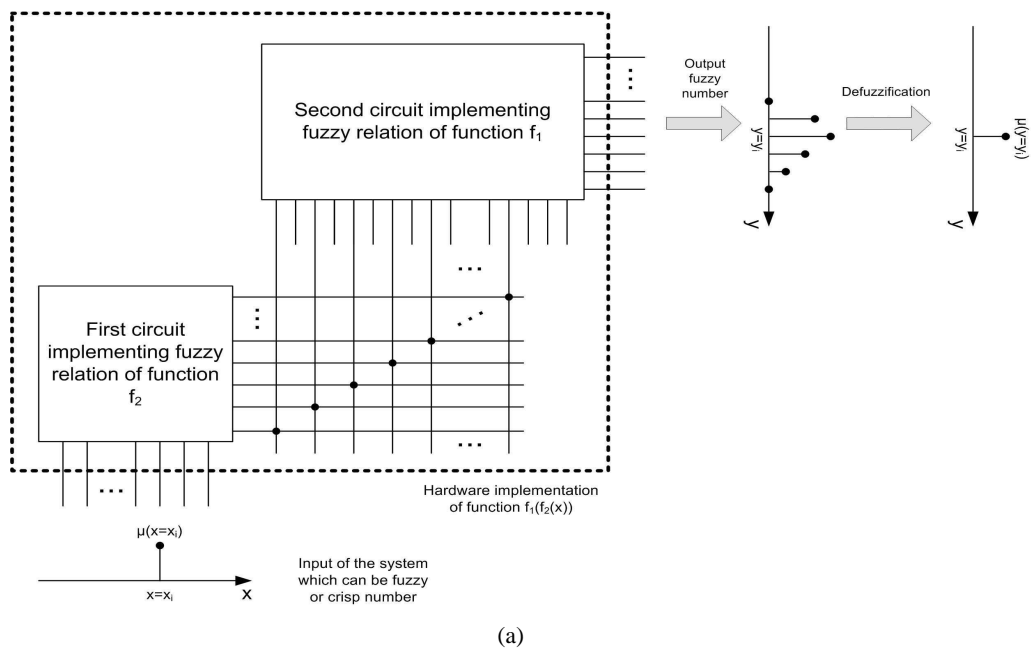}}
\vspace{0.14in} \subfigure[]{
\label{figcomb:b} 
\includegraphics[width=5.3in,height=2in]{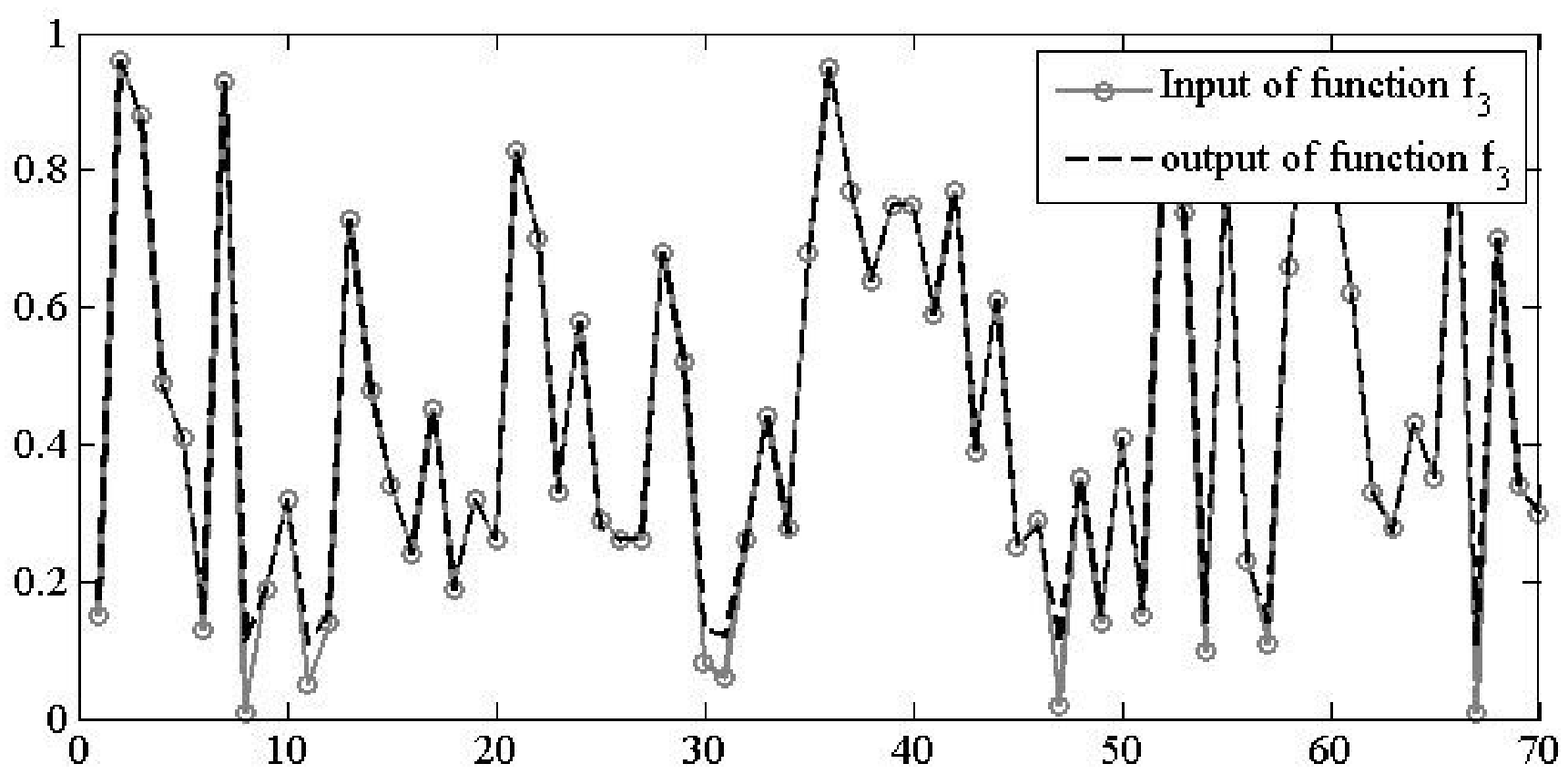}}
 \caption{(a) This figure shows that how simple circuits can be combined with each other to create complex functions.
 (b) Result of simulating function $f_3(x)=f_1(f_2(x))=x$ where this function is constructed by sequentially connecting two
 circuits implementing functions $f_1$ and $f_2$.}
\label{figcomb} 
\end{figure}

  Now, let's see how these two programmed
  circuits where each of them is a representative of one function can be combined with each other to create new functions.
  Without loss of generality, consider the creation of function
  $f_3(x)=f_1(f_2(x))=x$. To build this function, it is only sufficient to
  directly connect outputs of the circuit on which function $f_2$ is
  constructed to their corresponding inputs in the circuit which
  represents function $f_1$. This process is depicted schematically
  in Fig. \ref{figcomb:a}. Note that since the resolution of output variable in
  function $f_2$ may not be the same as the resolution of input
  variable in function $f_1$, in Fig. \ref{figcomb:a} output terminals of the first
  circuit is not connected sequentially to input terminals of the second
  circuit. Connecting two circuits representing functions $f_1$ and
  $f_2$ with fuzzy relations of Figs. \ref{figfunfafb:a} and
 \ref{figfunfafb:b} respectively in
  this way will create a new function $f_3(x)=f_1(f_2(x))=x$. Since $f_3(x)=x$ is a
  identity function its input and output should always be the
  same. Figure \ref{figcomb:b} shows the result of testing this circuit which is obtained by applying randomly
  generated crisp number to the input of the system and then
  plotting the result of the defuzzification of the output fuzzy
  number versus original input. This figure shows that the newly
  constructed function by merging two functions $f_1(x)=x^2$ and
  $f_2(x)=\sqrt{x}$ is a good approximation of function $f_3(x)=f_1(f_2(x))=x$.
  From the results of this simulation, it can be concluded that by sequentially connecting several circuits of
  Fig. \ref{fig3} where each of them represents
  different but simple function, any complex function can be
  reproduced. Here it is worth to mention that since in combinational circuits like
  the one shown in Fig. \ref{figcomb:a} output of one circuit directly connects to inputs of other circuit,
  input signals of one circuit will come from previous circuits. Actually, these are fuzzy
  numbers that propagate in the structure from one block to others and therefore no defuzzification process will
  be required in the entire system. Moreover, In this case
  there will be no need to any other auxiliary circuits to generate these
  signals since they are coming from other blocks.

\begin{figure}[!t]
\centering 
{
\includegraphics[width=6.8in,height=3.5in]{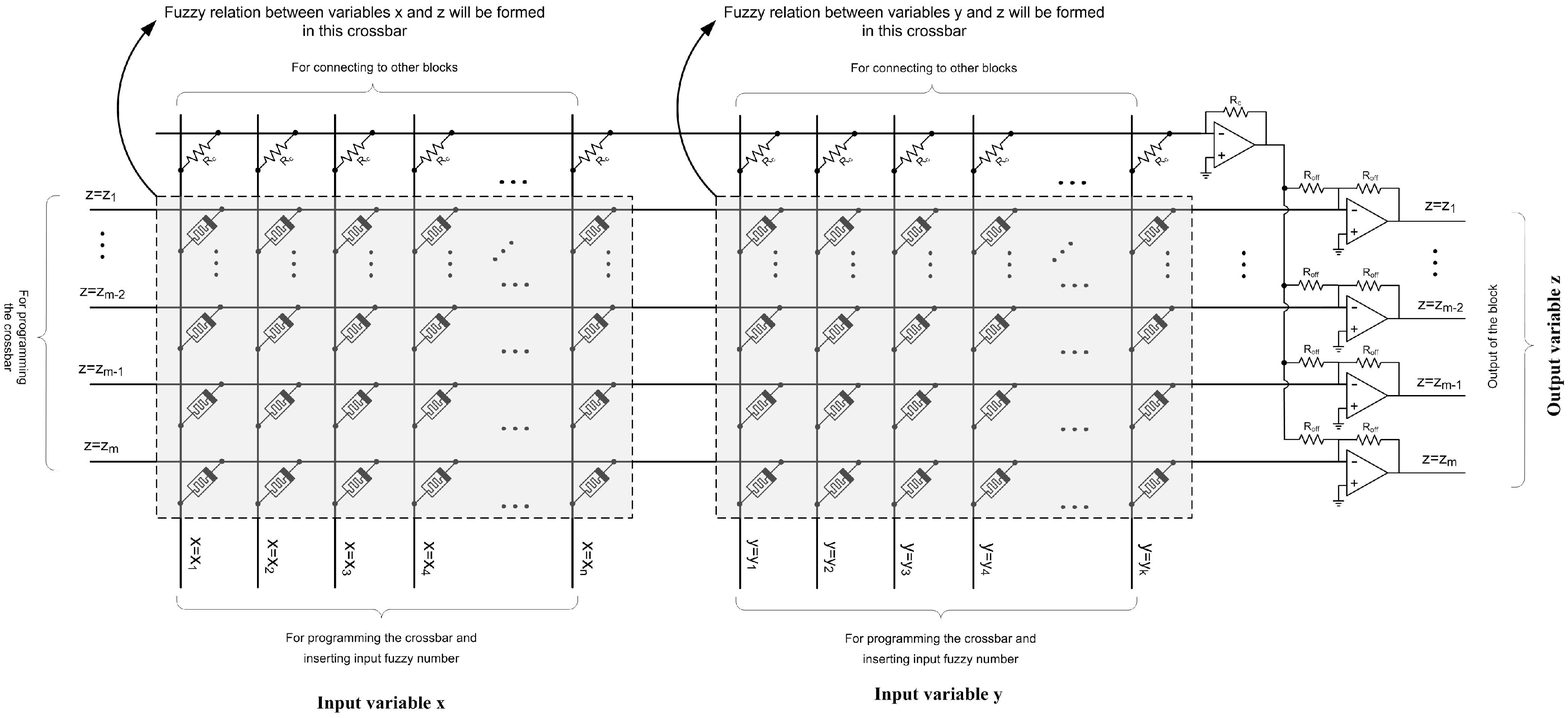}}
\caption{Structure of Fig. \ref{fig3} can be used to implement
multi-input functions. For this purpose, one can split input
terminals into several sections one for each of input variables.
Training of this structure will be the same az before but it should
be noted that in spite of the structure of Fig. \ref{fig3}, in this
structure several fuzzy relation will be formed on the crossbar.}
\label{figmultinpfun} 
\end{figure}

\begin{figure}[!t]
\centering \subfigure[]{
\label{fig161:a} 
\includegraphics[width=3.3in,height=2.4in]{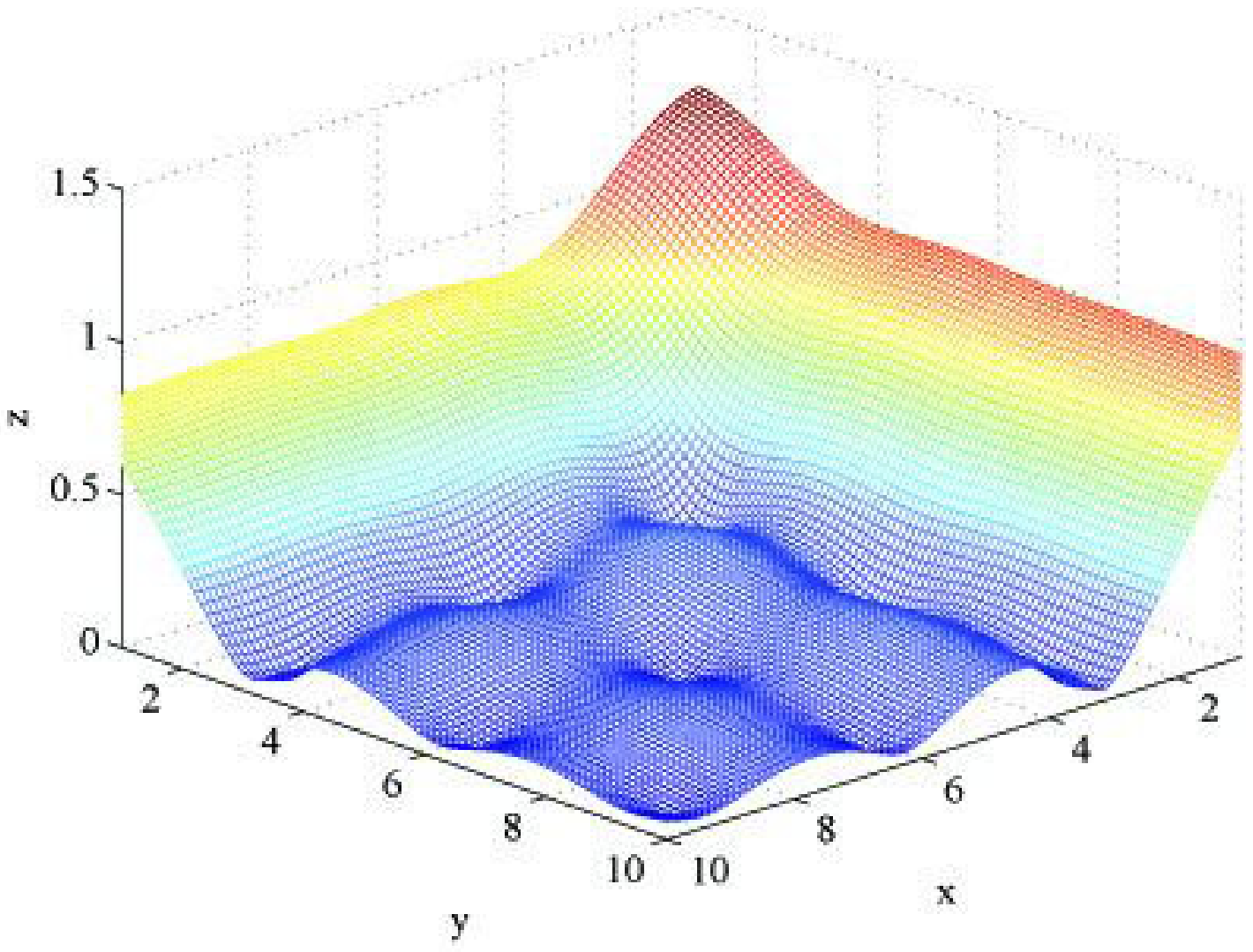}}
\vspace{0.14in} \subfigure[]{
\label{fig161:b} 
\includegraphics[width=3.3in,height=2.4in]{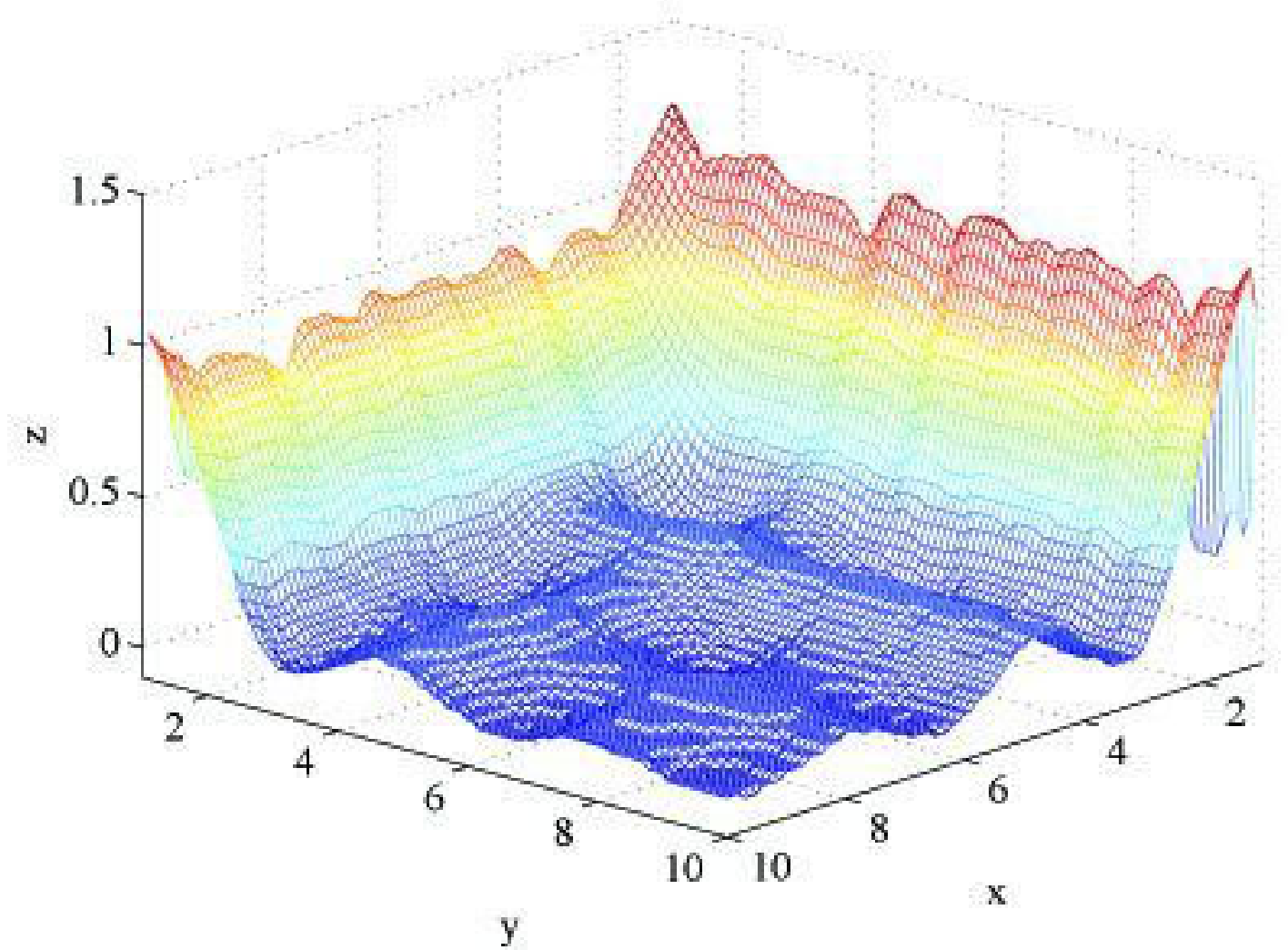}}
\vspace{0.14in} \subfigure[]{
\label{fig161:c} 
\includegraphics[width=3.3in,height=2.4in]{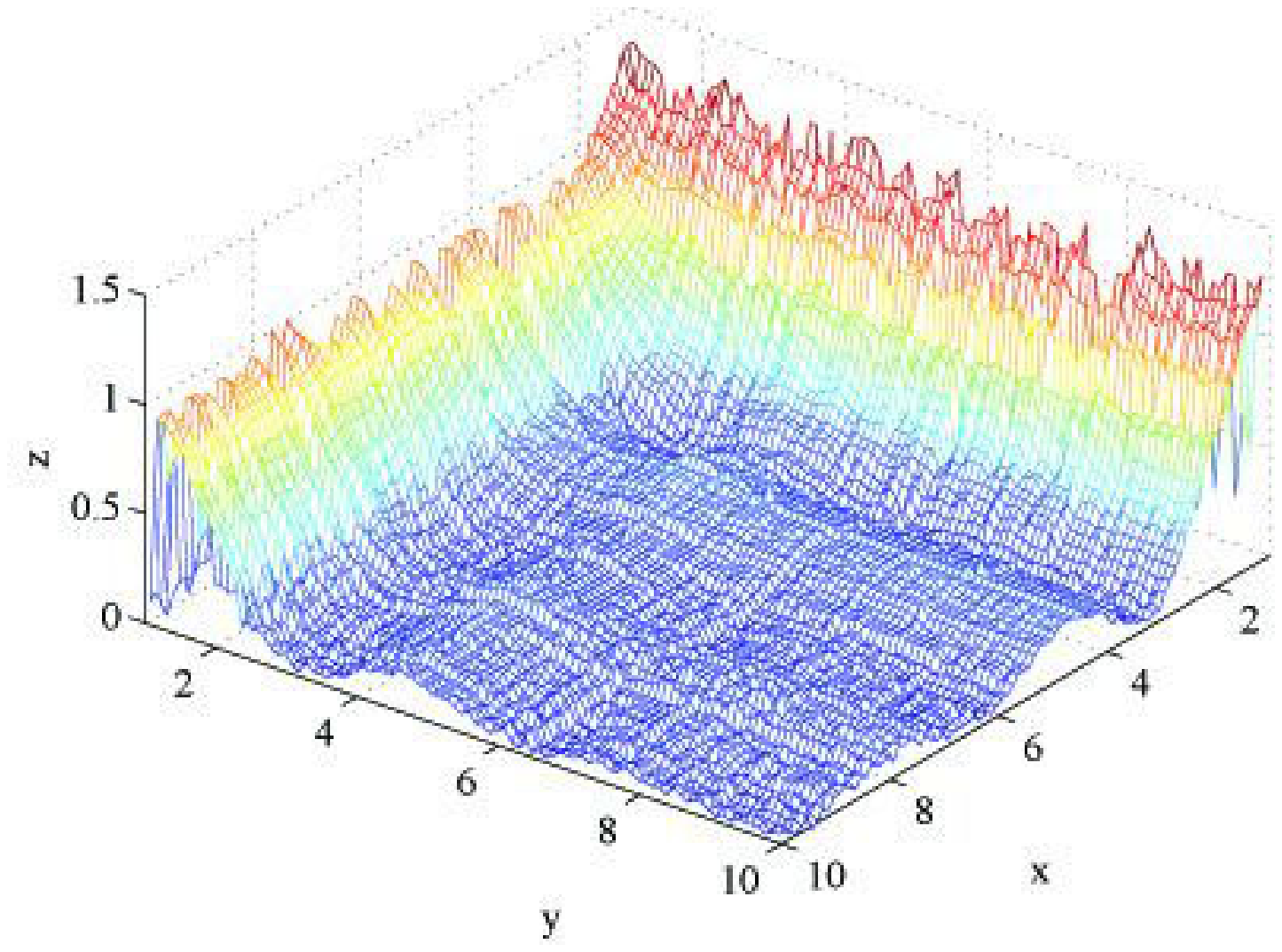}}
 \caption{This figure shows the ability of our proposed method in modeling 2-input functions. (a) graph of 2-input
 nonlinear function defined in Eq. \ref{eq161}. (b) Modeling result obtained by using 800 training data. (c) Modeling result
 obtained by using 800 training data while about half of the memristors of the crossbar are defective.}
\label{fig161} 
\end{figure}

  Functions with more than one input variable play an important roles in the construction of complex systems and inference methods.
  Note that functions like $AND$, $OR$, $MIN$ and $MAX$ belong to this category. Until
  now, we have deal only with 1-input functions in the form of $y=f(x)$ and our
  proposed structure in Fig. \ref{fig3} seems to be able to implement only these
  kinds of functions (because each binary fuzzy relation has one input and one output variable).
  In this part, we want to show how efficiently our proposed method and circuit can
  model multi-input functions. To use the
  structure of Fig. \ref{fig3} for modeling multi-input functions, there is no
  need to apply any changes to it. This is because of the fact that
  actually
  this structure in its current form is inherently implementing multi-input functions.
  To demonstrate this concept, note that if we assume that the
  concepts which are assigned to different input terminals of the circuit of Fig. \ref{fig3} (its vertical
  wires) be completely independent from each other, then each of
  these inputs can be considered as a distinct input variable. For example,
  assume that
  we want to implement a 2-input function $z=f(x,y)$ on the structure
  of Fig. \ref{fig3}. For this purpose, we can split the vertical wires of this circuit into
  two sections; wires in one section represent different values of
  one input variable and wires in other section represent different
  values of other input variable. This process is depicted in Fig.
  \ref{figmultinpfun}
  for better illustration. Note that by adding more vertical wire to
  the structure of Fig. \ref{fig3}, resolution and domain of each of these input variables
  can be simply increased. It is obvious that generalization of this
  method for the implementation of multi-input functions with any number of input
  variables
  is straightforward. Since we have not changed the circuit of Fig. \ref{fig3},
  learning of this circuit (creation of fuzzy
  relation) can be done the same as before. For this purpose, it is
  sufficient to apply fuzzy numbers of independent input
  variables to their corresponding columns of the crossbar and the fuzzy number of output
  variable to rows of the crossbar for $t_0$ second(s).
  Repeating this process for other fuzzy input-output training
  data will cause the fuzzy relation to be formed on the memristor
  crossbar of the circuit. However, in spite of previous tests, when implementing multi-input functions
  in this way instead of single fuzzy relation, several fuzzy relations will be formed on the crossbar.
  Actually, in this case one fuzzy relation per each input variable will be created on the crossbar as
  shown in Fig. \ref{figmultinpfun} as well. Each of these fuzzy relations specifies the overall behavior of output
  variable versus one of input variables. In this situation, structure's output fuzzy number for one given input sample
  will be computed from several intermediate output fuzzy numbers where each of them is obtained through one of these
  fuzzy relations based on the same method described in Section \ref{secdoinf}. To be precise, in fact as shown in Fig. \ref{figinfbb} as well, these fuzzy
  numbers are summed to each other to generate final output fuzzy number corresponding to given input. To illustrate this procedure
  better, we do
  several simulations. In the first simulation, ability of our
  proposed structure in modeling 2-input functions is investigated.
  The following nonlinear function is considered for this purpose:
\begin{equation}\label{eq161}
z=0.5\sqrt{2\left( \frac{\sin x}{x}\right)^2+3\left( \frac{\sin
y}{y}\right)^2},\qquad 1\leq x, y\leq10,
\end{equation}
where its graph is shown in Fig. \ref{fig161:a}. This modeling test
is conducted on the structure similar to the circuit depicted in
Fig. \ref{figmultinpfun} with 180 vertical wires (90 wires for each
of input variables) and 100 horizontal wires (for output variable).
Since input variables are bounded between 1 and 10 and there are 90
wires to cover this interval, minimum achievable resolution for
input variables by assuming that these wires are spread uniformly
over this interval will be:
\begin{eqnarray}\label{minresol}
&\ &\text{minimum achievable resolution for}\ x\ \text{and}\
y=\frac{\text{max}(x)-\text{min}(x)}{\text{number of reserved wires}}\nonumber\\
&=&\frac{\text{max}(y)-\text{min}(y)}{\text{number of reserved
wires}}=\frac{10-1}{90}=0.1,
\end{eqnarray}

\begin{figure}[!t]
\centering \subfigure[]{
\label{fuzxyzrel:a} 
\includegraphics[width=3.2in,height=2.2in]{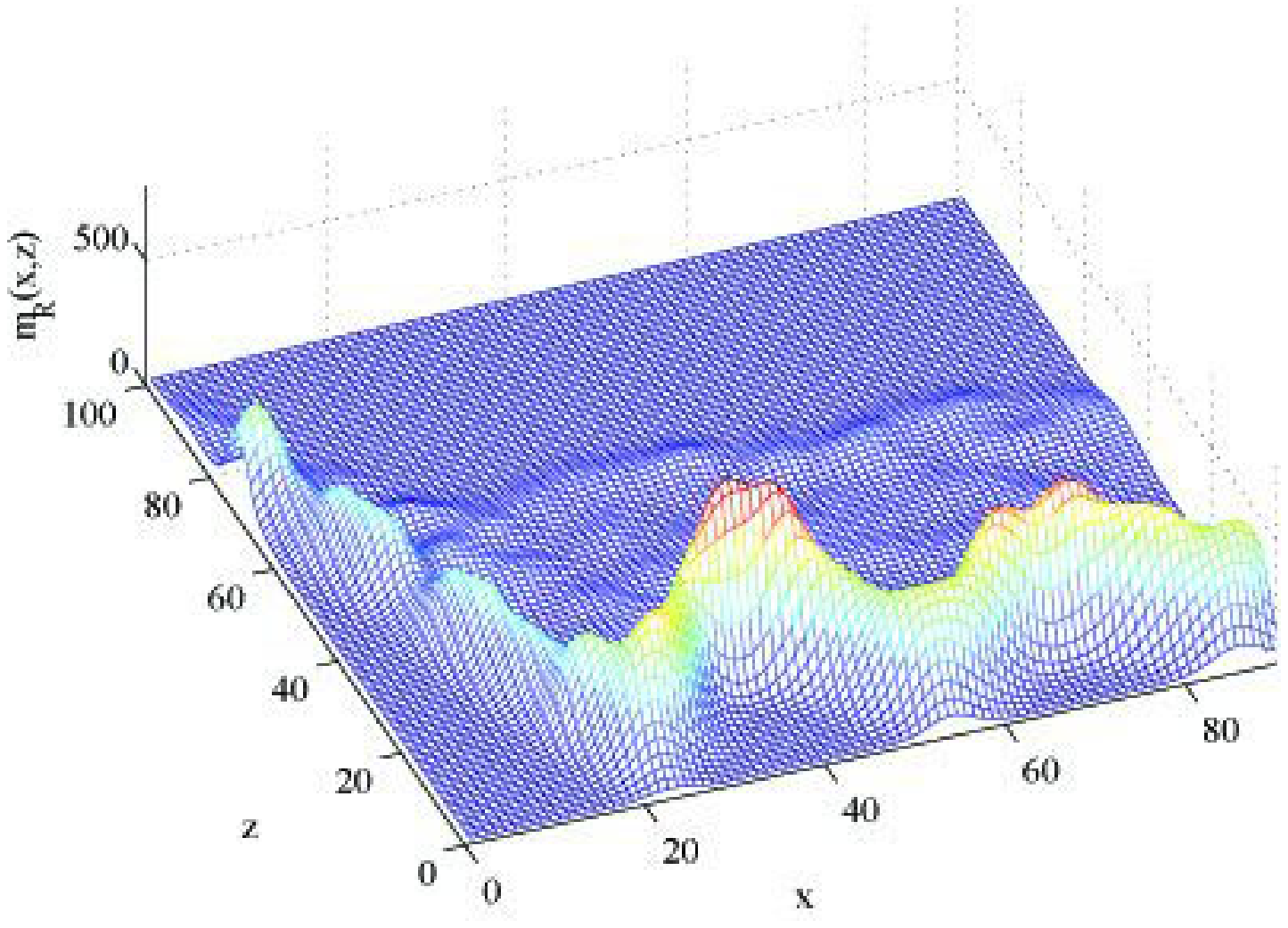}}
\vspace{0.14in} \subfigure[]{
\label{fuzxyzrel:b} 
\includegraphics[width=3.2in,height=2.2in]{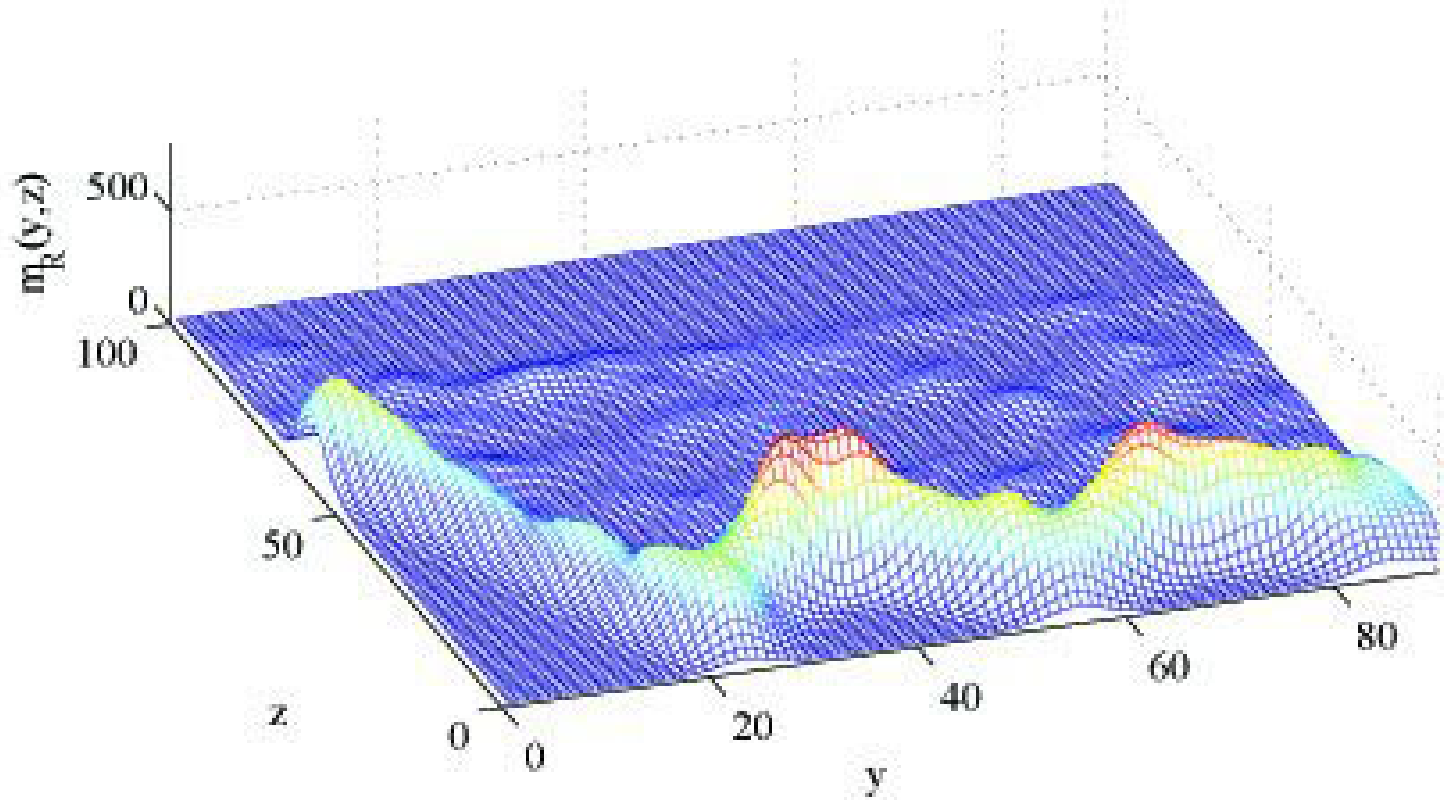}}
 \caption{Created fuzzy relations on the structure of Fig. \ref{figmultinpfun} during the modeling
 of 2-input nonlinear function defined in Eq. \ref{eq161}. (a) Created fuzzy relation between variables $x$ and $z$.
 (b) Created fuzzy relation between variables $y$ and $z$.}
\label{fuzxyzrel} 
\end{figure}

In a similar way, minimum achievable resolution for output variable
$z$ can be computed which becomes equal to $0.0112$. To create fuzzy
relations on the structure of Fig. \ref{figmultinpfun}, 800 fuzzy
input-output training data are used. Since no fuzzy input-output
data are available, we have to create them artificially. For this
purpose, at first 800 crisp input-output training data which are
uniformly distributed over input space are generated and then they
are converted to their corresponding fuzzy input-output training
data by using gaussian membership function. Application of these
training data to rows and columns of the crossbar of Fig.
\ref{figmultinpfun} will cause fuzzy relations to be created on the
crossbar. Figures \ref{fuzxyzrel:a} and \ref{fuzxyzrel:b} show these
formed fuzzy relations on the crossbar separately after this
training process. Now, similar to what is done in previous
simulation, {\it i.e.} by application of randomly generated crisp
data and defuzzification of produced output fuzzy number, the model
of function defined in Eq. \ref{eq161} can be reconstructed. The
result of this modeling test is presented in Fig. \ref{fig161:b}.
The Mean Square Error (MSE) between the target function and
constructed model is 0.021. This simulation clearly demonstrates the
efficiency of our proposed inference method in modeling multi-input
functions.

Nowadays, nano-scale devices and molecular electronics promise to
overcome the fundamental physical limitation of lithography-based
silicon VLSI technology \cite{Butts}. Furthermore, it has been
demonstrated that nano devices such as nanoscale crossbars can be
fabricated efficiently by using bottom-up self-assembly techniques
without relying on lithography to define the smallest feature size
\cite{Butts,Huang}. However, non-determinism in bottom-up
self-assembly chemical processes at molecular scale results in more
defects compared to highly controlled lithography-based
manufacturing processes currently used in CMOS technologies
\cite{Huang}. Therefore, defect tolerance is necessary for circuits
which are realized through the usage of nano-scale devices
\cite{Collier,DeHon}. In the next simulation, we will illustrate the
excellent fault tolerance capability of our proposed hardware.

Figure \ref{fig161:c} shows the result of modeling test obtained by
repeating the above simulation but this time with the hardware which
has some defects. In this test, we have assumed that approximately
half of the memristors in the memristor crossbar of the circuit of
Fig. \ref{figmultinpfun} is not working. This means that these
randomly chosen memristors have the memristance $R_{off}$ and their
memristance cannot be changed during the training process. In other
words, we have assumed that these faulty memristors behave
completely the same as a simple resistor with the resistance of
$R_{off}$ and nothing can be stored at these points. The training
set and the value of other parameters in this simulation are the
same as the previous test. The Mean Square Error (MSE) between the
target function and constructed model is 0.0281. Figure
\ref{fig161:c} indicates that although 50 percent of all of the
memristors in the crossbar are working incorrectly, our proposed
hardware still can model functions with acceptable accuracy.

\section{conclusion}
\label{conclusion} In this paper we proposed a new fuzzy inference
system and a simple method to create fuzzy relation based on input
and output fuzzy sets. In addition our proposed scheme has this
benefit that fuzzy relation can be formed in it based on fuzzy
training data and therefore offering learning capability. Since one
of the main problems of fuzzy systems relates to their efficient
hardware implementation, we have also designed simple memristor
crossbar-based circuit as a hardware implementation of our proposed
inference method. Simulation results show that this structure can
effectively be used to construct multi-input functions. Moreover, we
illustrated that samples of this memristor crossbar-based analog
circuit can be sequentially connected to each other to create a
cellular structure on which complex functions can be constructed.
Finally, according to the similarities between our proposed fuzzy
inference method and recently suggested memristor crossbar-based
circuits for constructing artificial neural networks, we showed that
if input and output signals of neural networks be of a kind of
confidence degree, then the computational task which is done in
conventional neural network will become equal to our proposed fuzzy
inference method with extra advantages.


\begin{thebibliography}{1}


\bibitem{anfis}
J.-S.R. Jang,
\newblock``Adaptive-Network-based Fuzzy Inference Systems,'' IEEE Transactions on Systems, Man ans Cybernetics, Vol. 23, pp. 665--685, 1993.

\bibitem{RulNet}
N. N. Tschichold-Gurman,
\newblock``The Neural Network Model RuleNet and its Application to Mobile Robot Navigation,'' Fuzzy Sets and Systems, Vol. 85, pp. 287--303, 1997.

\bibitem{GARIC}
H. R. Berenji, and P. Khedkar,
\newblock``Learning and Tuning Fuzzy Logic Controllers through Reinforcements,'' IEEE Transactions on Neural Networks, Vol. 3, pp. 724--740, 1992.

\bibitem{williams}
D.B. Strukov, G.S. Snider, D.R. Stewart and R.S. Williams,
\newblock``The missing memristor found,'' Nature, vol. 453, pp. 80--83, 1 May 2008.

\bibitem{Moneta}
M. Versace, and B. Chandler,
\newblock``The Brain of a New Machine,'' IEEE Spectrum,  vol. 47, No. 12, pp. 30--37, 2010.



\bibitem{leszek}
L. Rutkowski,
\newblock``Flexible Neuro-Fuzzy Systems: Structures, Learning and Performance Evaluation,'' Kluwer Academic Publishers, 2004.


\bibitem{zadehimplication}
L. A. Zadeh,
\newblock``The concept of a linguistic variable and its application to approximate reasoning,''
 Information Sciences, Vol. 8, No. 3, pp. 199--249, 1975.


\bibitem{fodor}
J. C. Fodor,
\newblock``On fuzzy implication operators,''
 Fuzzy sets and systems, Vol. 24, pp. 293--300, 1991.

\bibitem{dubois}
D. Dubois, J. Lang, and H. Prade,
\newblock``Fuzzy sets in approximate reasoning,''
 Fuzzy sets and systems, Vol. 40, No. 1, pp. 143--244, 5 March 1991.


\bibitem{shouraki1}
S.B. Shouraki,
\newblock``A Novel Fuzzy Approach to Modeling and Control and
its Hardware Implementation Based on Brain Functionality and
Specifications,'' Ph.D. dissertation, The Univ. of
Electro-Communications, Chofu, Japan, March 2000.


\bibitem{murakami}
M. Murakami and N. Honda,
\newblock``A study on the modeling ability of the IDS
method: A soft computing technique using pattern-based information
processing,'' International Journal of Approximate Reasoning, vol.
45, pp. 470--487, 2007.


\bibitem{mamdani}
E. H. Mamdani, and S. Assilian,
\newblock``An Experiment in Linguistic Synthesis with a Fuzzy Logic Controller,'' International Journal
of Human-Computer Studies, vol. 51, No. 2, pp. 135--147, 1999.


\bibitem{sugeno}
T. Takagi, and M. Sugeno,
\newblock``Fuzzy Identification of Systems and its Application to Modeling and Control,'' IEEE Transaction on
Systems, Man and Cybernetics, vol. 15, pp. 116--132, 1985.




\bibitem{Chua}
L.O. Chua,
\newblock``Memristor - the missing circuit element,''
IEEE Trans. on Circuit Theory, vol. CT-18, no. 5, pp. 507--519,
1971.


\bibitem{waser}
R. Waser, and M. Aono,
\newblock``Nanoionics-based resitive switching memories,''
Nature Materials 6, vol. pp. 833--840, 2007.


\bibitem{pershin}
Y.V. Pershin, S.L. Fontaine, and M.D. Ventra,
\newblock``Memristive model of amoeba's learning,''
 Phys. Rev. E, vol. 80, p. 021926, 2009.


\bibitem{Shin}
S. Shin, K. Kim, and S.M. Kang,
\newblock``Memristor-based fine resolution resistance and its applications,''
 ICCCAS 2009, July 2009.


\bibitem{perShin22}
Y. V. pershin, and M.D. Ventra,
\newblock``Practical Approach to Programmable Analog Circuits With Memristors,''
 IEEE Transactions on Circuits and Systems I: Regular Paper, Vol. 57, No. 8, pp. 1857--1864, Aug. 2010.



\bibitem{farelsevmemarith}
F. Merrikh-bayat, and S. B. Shouraki,
\newblock``Memristor-based circuits for performing basic arithmetic operations,''
 Procedia-Computer Science Journal, Vol. 3, pp. 128--132, 2011.


\bibitem{farIEEE}
F. Merrikh-bayat, and S. B. Shouraki,
\newblock``Memristor Crossbar-based Hardware Implementation of IDS Method,''
 Submitted to IEEE Transaction on Fuzzy Systems.

\bibitem{kuekes}
P. Kuekes,
\newblock``Material Implication: digital logic with memristors,''
Memristor and Memristive Systems Syymposium, 21 November  2008.

\bibitem{Mouttet1}
B.L. Mouttet,
\newblock``Proposal for Memristors in Signal Processing,''
 Nano-Net Conference, Vol. 3, pp. 11--13, Sept. 2008.

\bibitem{Mouttet2}
F. Merrikh-Bayat, and S. B. Shouraki,
\newblock``Memristor Crossbar-based Hardware Implementation of Sign-Sign LMS Adaptive Filter,''
 Analog Integrated Circuits and Signal Processing, DOI: 10.1007/s10470-010-9523-3.


\bibitem{williams2}
J.J. Yang, M.D. Pickett, X. Li, D.A. Ohlberg, D.R. Steward, and R.S.
Williams,
\newblock``Memristive switching mechanism for metal/oxide/metal nanodevices,''
Nature Nanotechnology 3, pp. 429-433, 2008.


\bibitem{crossbar}
P. J. Kuekes, D. R. Stewart, and R. S. Williams,
\newblock``The crossbar latch: Logic value storage, restoration, and inversion in crossbar circuits,''
 Journal of Applied Physics, Vol. 97, No. 3, 2005.


\bibitem{memnonvol}
N. Gergel-Hackett, B. Hamadani, B. Dunlap, J. Suehle, C. Richter, C.
Hacker, and D. Gundlach,
\newblock``A Flexible Solution-Processed Memristor,''
 IEEE ELECTRON DEVICE LETTERS, VOL. 30, NO. 7, pp. 706--708, JULY 2009.


\bibitem{memrismodel}
W. Wang, Q. Yu, C. Xu, and Y. Cui,
\newblock``Study of Filter Characteristics Based on PWL Model,''
 International Conference on Communications, Circuits and Systems, pp. 969--973, 2009.


\bibitem{weilu}
S. H. Jo, T. Chang, I. Ebong, B. Bhavitavya, P. Mazumder and W. Lu,
\newblock``Nanoscale Memristor Device as Synapse in Neuromorphic Systems,''
 Nano Letter, 10, pp. 1297--1301, 2010.

\bibitem{Chabi}
D. Chabi, and J.-O. Klein,
\newblock``Hight Fault Tolerance in Neural Crossbar,''
 International Conference on Design and Technology of Integrated Systems in Nanoscale Era, pp. 1--6, 2010.


\bibitem{Cantley}
K. Cantley, A. Subramaniam, H. Stiegler, R. Chapman and E. Vogel,
\newblock``Hebbian Learning in Spiking Neural Networks with Nano-Crystalline Silicon TFTs and Memristive Synapses,''
 Accepted in IEEE Transactions on Nanotechnology, 2011.



\bibitem{afifi}
A. Afifi, A. Ayatollahi and F. Raissi,
\newblock``Implementation of Biologically Plausible Spiking Neural Network Models on the Memristor Crossbar-based CMOS/Nano Circuits,''
 European Conference on Circuit Theory and Design, pp. 563--566, 2009.


\bibitem{Carrasco}
J. A. Carrasco, C. Zamarreno-Ramos, T. Serrano-Gotarredona, and B.
Linares-Barranco,
\newblock``On Neuromorphic Spiking Architectures for Asynchronous STDP Memristive Systems,''
 IEEE International Symposium on Circuits and Systems, pp. 1659--1662, 2010.


\bibitem{Snider}
G. Snider,
\newblock``Spike-Timing-Dependent Learning in Memristive Nanodevices,''
 IEEE/ACM International Symposium on Nanoscale Architectures, pp. 85--92, Anaheim, CA, 2008.

\bibitem{faset}
L. V. Fausett,
\newblock``Fundamentals of Neural Networks: Architectures, Algorithms and Applications,''
 Prentice Hall, 1993.


\bibitem{Hebb}
D. O. Hebb,
\newblock``The Organization of Behavior; A Neuropsychological Theory,''
 Wiley-Interscience,New York, 1949.


\bibitem{farmemproblem}
F. Merrikh-bayat, and S. B. Shouraki,
\newblock``Battleneck of using single memristor as a synapse and its solution,''
 Submitted to Neural Processing Letters, Springer.


\bibitem{Biolek}
D. Biolek, Z. Biolek and V. Biolkova,
\newblock``SPICE Modeling of Memristive, Memcapacitative
and Meminductive Systems,''
 European Conference on Circuit Theory and Design (ECCTD2009), pp. 249--252, Antalya, 23--27 Augost 2009.


\bibitem{Butts}
M. Butts, A. DeHon, and S. C. Goldstein,
\newblock``Molecular Eletronics: Devices, Systems and Tools for Gigagate, Gigabit
Chips,''
 Proc. International Conference on Computer-Aided Design, pp. 443--440, 2002.


\bibitem{Huang}
J. Huang, M. B. Tahoori, and F. Lombardi,
\newblock``On the defect tolerance of nano-scale two-dimensional crossbars,''
 19th IEEE International Symposium on Defect and Fault Tolerance in VLSI Systems, pp. 96--104, Oct. 2004.


\bibitem{Collier}
C. P. Collier, E. W. Wong, M. Belohradsky, F. M. Raymo, J. F.
Stoddart, P. J. Kuekes, R. S. Williams, and J. R. Heath,
\newblock``Electronically Configurable Molecular-Based Logic Gates,''
 IEEE Trans. on Nanotechnology, Science, vol. 285, pp. 391--394, 1999.

\bibitem{DeHon}
A. DeHon,
\newblock``Array-Based Architecture for FET-Based, Nanoscale Electronics,''
 IEEE Trans. on Nanotechnology, vol. 2, No. 1, pp. 23--32, 2003.


\end{thebibliography}
\end{document}